\documentclass[11pt, english]{article}
\usepackage{babel}
\usepackage[latin1]{inputenc}
\usepackage[T1]{fontenc}
\usepackage{amsmath}
\usepackage{graphicx}
\usepackage{amssymb}
\usepackage{amsthm}
\usepackage{subcaption}
\usepackage{bbm}
\usepackage{mathtools}

\usepackage{algorithmic}
\usepackage{courier} 
\usepackage{algorithm}
\graphicspath{{figures/}}

\usepackage[round]{natbib}

\usepackage{chngcntr}
\usepackage{apptools}
\AtAppendix{\counterwithin{thm}{section}}

\def\^{\widehat}

\def\phi{\varphi}

\numberwithin{equation}{section}

\renewcommand{\phi}{\varphi}

\def\~{\widetilde}
\def\^{\widehat}
\newcommand{\ee}{{\rm e}\hspace{1pt}}

\newcommand{\dd}{\hspace{1pt}{\rm d}\hspace{0.5pt}}
\newcommand{\abs}[1]{\left| #1 \right|}
\newcommand{\wt}[1]{\widetilde{#1}}
\newcommand{\deltas}[1]{\delta_{#1}(s)}
\newcommand{\deltat}[1]{\delta_{#1}(t)}

\newcommand{\veps}{\varepsilon}

\newtheorem{thm}{Theorem}

\newtheorem{lem}[thm]{Lemma}
\newtheorem{cor}[thm]{Corollary}
\newtheorem{defn}[thm]{Definition}

\newtheorem{remark}[thm]{\textit{Remark}}

\title{Tight Differential Privacy for Discrete-Valued Mechanisms \\ and for the Subsampled Gaussian Mechanism Using FFT}

\author{Antti Koskela$^{1}$, Joonas J\"alk\"o$^{2}$, Lukas Prediger$^{2}$ and Antti Honkela$^1$ \vspace{5mm} \\ 
$^1$ Helsinki Institute for Information Technology HIIT,\\
    Department of Computer Science, University of Helsinki, Finland \\
  $^2$ Helsinki Institute for Information Technology HIIT,\\
    Department of Computer Science, Aalto University, Finland }

    \date{}

\begin{document}
	
\maketitle

\abstract{

We propose a numerical accountant for evaluating the tight $(\varepsilon,\delta)$-privacy loss for algorithms with discrete one dimensional output. The method is based on the privacy loss distribution formalism and it uses the recently introduced fast Fourier transform based accounting technique. We carry out an error analysis of the method in terms of moment bounds of the privacy loss distribution which leads to rigorous lower and upper bounds for the true $(\varepsilon,\delta)$-values. As an application, we present a novel approach to accurate privacy accounting of the subsampled Gaussian mechanism. This completes the previously proposed analysis by giving strict lower and upper bounds for the privacy parameters. We demonstrate the performance of the accountant on the binomial mechanism and show that our approach allows decreasing noise variance up to 75 percent at equal privacy compared to existing bounds in the literature. We also illustrate how to compute tight bounds for the exponential mechanism applied to counting queries.

}

\section{Introduction}

Differential privacy (DP)~\citep{dwork_et_al_2006} has been established
as the standard approach for privacy-preserving machine learning.  As
DP algorithms have grown increasingly complex, accurately bounding the
compound privacy loss has become more challenging as well.  The
moments accountant~\citep{Abadi2016} represented a major
breakthrough in the accuracy of bounding the privacy loss in
compositions of subsampled Gaussian mechanisms that are commonly used
in DP stochastic gradient descent (DP-SGD).  This has further been
refined through the general development of R\'enyi differential
privacy (RDP)~\citep{mironov2017} as well as tighter RDP bounds for
subsampled mechanisms~\citep{balle2018subsampling,wang2019,zhu2019,mironov2019}. 
RDP enables tight analysis for compositions of Gaussian
mechanisms, but this may be difficult for other mechanisms.
Moreover, conversion of RDP guarantees back to more commonly used
$(\varepsilon, \delta)$-guarantees is lossy.

In this work, we focus on an alternative approach 
based on the privacy loss distribution (PLD)
formalism introduced by~\citet{sommer2019privacy}. This work directly extends the
recent Fourier accountant by~\citet{koskela2020} to discrete mechanisms.
We provide a rigorous error analysis  which leads to strict 
$(\veps,\delta)$-bounds.
This analysis is further used to obtain strict bounds 
for the subsampled Gaussian mechanism. 

The need to consider discrete mechanisms for rigorous DP on
finite-precision computers was first pointed out by~\citet{mironov2012}.~\citet{agarwal2018}
implement a communication efficient binomial mechanism cpSGD
for 
neural network training which however cannot handle
compositions. 
\citet{agarwal2018} and~\citet{kairouz2019} note the need for a privacy accountant for the
binomial mechanism as an important open problem, which we solve in this paper
for the case where gradients are replaced with a sign approximation.

The outline of the paper is as follows.
In Sections 2 and 3 we give the basic definitions and describe the PLD formalism used for our accountant.
In Section 4 we describe the algorithm based on the fast Fourier transform (FFT) and in Section 5 we provide an error analysis. 
Section 6 concludes with experiments illustrating the efficiency and accuracy of the method.

	 Implementation of the methods is available in Github\footnote{https://github.com/DPBayes/PLD-Accountant/}.

%
\paragraph{Our Contribution.}
\hspace{-3mm} We extend the work by~\citet{koskela2020} which considered an FFT based method for approximating
the tight $(\veps,\delta)$-DP guarantees of the subsampled Gaussian mechanism, however without strict lower and upper bounds. The main contributions of this work are:
\begin{itemize}
\item  A framework for computing tight $(\veps,\delta)$-DP guarantees of discrete-valued mechanisms.
\item  An error analysis of the proposed method using moment bounds of the mechanism at hand, which leads to strict lower and $(\veps,\delta)$-upper bounds. 
\item  Accurate lower and upper bounds for $(\veps,\delta)$-DP of the subsampled Gaussian mechanism.
\end{itemize}

\section{Differential Privacy}

We first recall some basic definitions of DP~\citep{dwork_et_al_2006}. We use the
following notation. An input data set containing $N$ data points is denoted as $X = (x_1,\ldots,x_N)
\in \mathcal{X}^N$, where $x_i \in \mathcal{X}$, $1 \leq i \leq N $.

\begin{defn} \label{def:adjacency}
	We say two data sets $X$ and $Y$ are neighbours in remove/add relation if we get 
	one by removing/adding an element from/to the other and denote this with $\sim_R$.
	We say $X$ and $Y$ are neighbours in substitute relation if we get one by substituting
	one element in the other. We denote this with $\sim_S$.
\end{defn}

\begin{defn} \label{def:dp}
	Let $\varepsilon > 0$ and $\delta \in [0,1]$. Let $\sim$ define a neighbouring relation.
	Mechanism $\mathcal{M} \, : \, \mathcal{X}^N \rightarrow \mathcal{R}$ is $(\varepsilon,\delta,\sim)$-DP 
	if for every $X \sim Y$
	and every measurable set $E \subset \mathcal{R}$ we have that
	$$
		\mathrm{Pr}( \mathcal{M}(X) \in E ) \leq \ee^\varepsilon \mathrm{Pr} (\mathcal{M}(Y) \in E ) + \delta.
	$$
	When the relation is clear from context or irrelevant, we will abbreviate it as $(\veps, \delta)$-DP. 
	We call $\mathcal{M}$ tightly $(\veps,\delta,\sim)$-DP, if there does not exist $\delta' < \delta$
	such that $\mathcal{M}$ is $(\veps,\delta',\sim)$-DP.
\end{defn}

\section{Privacy Loss Distribution} \label{sec:pld}

We first introduce the basic tool for obtaining tight privacy bounds: the privacy loss distribution (PLD).
The results in Subsection~\ref{subsec:pld} are reformulations of the results given by~\citet{meiser2018tight} and~\citet{sommer2019privacy}.
Proofs of the results of this section are given in the supplementary material. 

\subsection{Privacy Loss Distribution}  \label{subsec:pld}

We consider discrete-valued one-dimensional mechanisms $\mathcal{M}$ which can be seen as mappings from $\mathcal{X}^N$
to the set of discrete-valued random variables.
The \emph{generalised probability density functions} of $\mathcal{M}(X)$ and $\mathcal{M}(Y)$, denoted $f_X(t)$ and $f_Y(t)$, respectively, are given by
\begin{equation} \label{eq:delta_sum}
	\begin{aligned}
		f_X(t) &= \sum\nolimits_i a_{X,i} \cdot \deltat{t_{X,i}}, \\
		f_Y(t) &= \sum\nolimits_i a_{Y,i} \cdot \deltat{t_{Y,i}},
	\end{aligned}
\end{equation}
where $\delta_t( \cdot )$, $t \in \mathbb{R}$, 
denotes the Dirac delta function centered at $t$, and $t_{X,i},t_{Y,i} \in \mathbb{R}$ and $a_{X,i},a_{Y,i} \geq 0$.
Equivalently, \eqref{eq:delta_sum} means that for all $i$,
\begin{equation*} 
	\begin{aligned}
		\mathbb{P}( \mathcal{M}(X) &=  t_{X,i}  ) = a_{X,i}, \\
		\mathbb{P}( \mathcal{M}(Y) &=  t_{Y,i}  ) = a_{Y,i}.
	\end{aligned}
\end{equation*}
Thus, we have that
\begin{equation*} 
	\begin{aligned}
\int\nolimits_{-\infty}^\infty f_X(t) \, \dd t &= \sum\nolimits_i a_{X,i} = 1, \\
\int\nolimits_{-\infty}^\infty f_Y(t) \, \dd t &= \sum\nolimits_i a_{Y,i} = 1.
	\end{aligned}
\end{equation*} 
If $g$ is a function such that $g(X)$ determines a random variable, then
\begin{equation} \label{eq:gpdf} 
	\begin{aligned}
\mathbb{E}_{s \sim X} [ g(s) ] &= \int\nolimits_{-\infty}^\infty  g(t) f_X(t) \, \dd t  \\
&= \sum\nolimits_i a_{X,i} \cdot g(t_{X,i}).
	\end{aligned}
\end{equation}
More generally, we define integrals over generalised probability density functions as in \eqref{eq:gpdf}.
We prefer using the integral notation 
as it simplifies the analysis.


We define the privacy loss distribution as follows.
\begin{defn} \label{def:pld}
Let $\mathcal{M} \, : \, \mathcal{X}^N \rightarrow \mathcal{R}$, $\mathcal{R} \subset \mathbb{R}$, be a discrete-valued randomised mechanism
and let $f_X(t)$ and $f_Y(t)$ be probability density functions of the form \eqref{eq:delta_sum}.
We define the privacy loss distribution $\omega_{X/Y}$ as   
\begin{equation} \label{eq:omega_pld}
	\begin{aligned}
	\omega_{X/Y}(s) &= \sum\nolimits_{{t_{X,i} = t_{Y,j} }}   a_{X,i} \cdot \deltas{s_i},
	\end{aligned}
\end{equation}
where $s_i = \log \left( \tfrac{a_{X,i}}{a_{Y,j}} \right)$.
\end{defn}
Notice that this definition differs slightly from the one given by~\citet[Def. 4.2]{sommer2019privacy}:
we do not include the symbol $\infty$ in $\omega$. Thus, if $f_X(t)$ and $f_Y(t)$ do not have equal supports,
we have $\int_\mathbb{R} \omega_{X/Y}(s) \, \dd s < 1$. 
This situation is included in our Lemma~\ref{eq:max_eq1} and Theorem~\ref{thm:integral}, and
the analysis of Section~\ref{sec:err_est} also applies then. 
We remark that Def.~\ref{def:pld} is related to the KL divergence, as
$\mathrm{KL}(f_X || f_Y) = \mathbb{E}[\, \omega_{X/Y}]  = \int_{-\infty}^\infty s \cdot \omega_{X/Y}(s) \, \dd s$
in case $f_X$ and $f_Y$ have equal supports.


Evaluating $(\veps,\delta)$-bounds using the PLD formalism is essentially based on a result (Supplements) which
states that the mechanism $\mathcal{M}$ is tightly $(\veps,\delta)$-DP with
\begin{equation} \label{eq:max_eq1}
	\begin{aligned}
\delta(\veps) &= \max_{X \sim Y} \Bigg\{ \int\nolimits_\mathbb{R}  \max \{  f_X(t) - \ee^\veps f_Y(t) ,0  \} \, \dd t,  \\
 & \int\nolimits_\mathbb{R}  \max \{  f_Y(t) - \ee^\veps f_X(t) ,0  \} \, \dd t \Bigg\}.
	\end{aligned}
\end{equation}
This relation holds for both continuous and discrete output mechanisms, and a more general version of this result
using so called $f$-divergences is given by~\citet{barthe2013beyond}. 
In case $f_X$ and $f_Y$ are generalised probability density functions of the form \eqref{eq:delta_sum}, i.e.,
$$
f_X(t) - \ee^\veps f_Y(t) = \sum\nolimits_i c_i \cdot \deltat{t_i}
$$
for some coefficients $c_i,t_i \in \mathbb{R}$, then in \eqref{eq:max_eq1} we denote
$$
\max \{  f_X(t) - \ee^\veps f_Y(t) ,0  \}  = \sum\nolimits_i \max \{c_i,0\} \cdot \deltat{t_i}.
$$
For the discrete-valued mechanisms, the relation \eqref{eq:max_eq1} was originally given by~\citet[Lemmas 5 and 10]{sommer2019privacy}.
Assuming the PLD distribution is of the form \eqref{eq:omega_pld}, 
the relation \eqref{eq:max_eq1} directly gives the following representation 
for $\delta(\veps)$.

\begin{lem}  \label{lem:maxrepr1}
$\mathcal{M}$  is tightly $(\varepsilon,\delta)$-DP for 
$$
\delta(\veps) = \max_{X \sim Y} \, \{ \delta_{X/Y}(\veps), \delta_{Y/X}(\veps) \},
$$
where
\begin{equation} \label{eq:delta_inf}
	\begin{aligned}
		& \delta_{X/Y}(\veps) = \delta_{X/Y}(\infty) + \int_\veps^\infty  (1-\ee^{\veps - s}) \, \omega_{X/Y}(s)   \, \dd s, \\
		& \delta_{X/Y}(\infty) = \\
		& \sum\nolimits_{ \{ t_i \, : \, \mathbb{P}( \mathcal{M}(X) = t_i) > 0, \, \mathbb{P}( \mathcal{M}(Y) = t_i) = 0 \} } 
		\mathbb{P}( \mathcal{M}(X) = t_i),
	\end{aligned}
\end{equation}
and similarly for $\delta_{Y/X}(\veps)$.
\end{lem}

We remark that finding the outputs $\mathcal{M}(X)$ and $\mathcal{M}(Y)$ that give the maximum $\delta(\varepsilon)$ 
is application specific and has to be carried out individually for each case, similarly as, e.g., in the case of 
RDP~\citep{mironov2017}.

\subsection{Example: The Randomised Response}

To illustrate the formalism described above, consider the randomised response mechanism~\citep{warner1965} 
which is described as follows.
Suppose $F$ is a function $F \, : \, \mathcal{X} \rightarrow \{0,1\}$.
Define the randomised mechanism $\mathcal{M}$ for input $X \in \mathcal{X}$ by
$$
\mathcal{M}(X) = \begin{cases}
	F(X), &\text{ with probability } p \\
	1-F(X), &\text{ with probability } 1-p,
\end{cases}
$$
where $0<p<1$.  The mechanism is $\veps$-DP for $\veps=\log \tfrac{p}{1-p}$~\citep{DworkRoth}.
Let $X \sim Y$ and let $F(X)=1$ and $F(Y)=0$.
As these are the only possible outputs, $X$ and $Y$ 
represent the worst case in Lemma~\ref{lem:maxrepr1} and give the tight $\delta(\veps)$.
We see that the density functions of $\mathcal{M}(X)$ and $\mathcal{M}(Y)$ are given by
\begin{equation*}
	\begin{aligned}
		f_X(t) &= p \cdot \deltat{1} + (1-p) \cdot \deltat{0}, \\
		f_Y(t) &= (1-p) \cdot \deltat{1} + p \cdot \deltat{0}.
	\end{aligned}
\end{equation*}
From \eqref{eq:omega_pld} we see that 
\begin{equation*}
	\begin{aligned}
\omega_{X/Y}(s) &= p \cdot \deltas{c_p} + (1-p) \cdot \deltas{-c_p}, \\
\omega_{Y/X}(s) &= p \cdot \deltas{-c_p} + (1-p) \cdot \deltas{c_p},
	\end{aligned}
\end{equation*}
where $c_p = \log \tfrac{p}{1-p}.$ 
Assume $\tfrac{1}{2} < p < 1$. Then by Lemma~\ref{lem:maxrepr1} we see that
$$
\delta(\veps) = \begin{cases}
	p \; (1- \ee^{\veps-c_p}), &\text{if  $\veps \leq c_p$ }\\
    0, 	 &\text{else.}
\end{cases}
$$
As $\veps \rightarrow^- c_p$, we see that  $\delta \rightarrow 0$ as expected.

\subsection{Tight $(\veps,\delta)$-Bounds for Compositions}

Let $X$ and $Y$ be random variables described by generalised probability density functions
$f_X$ and $f_Y$ of the form \eqref{eq:delta_sum}.
We define the convolution $f_X * f_Y$ as
\begin{equation*}
	(f_X * f_Y )(t) =  \sum\nolimits_{i,j} a_{X,i} \, a_{Y,j} \cdot \deltat{t_{X,i} + t_{Y,j}}.  
\end{equation*}
Notice that $f_X * f_Y$ describes the probability density of the random variable $X + Y$. 
The following theorem shows that the tight $(\veps,\delta)$-bounds for compositions 
of non-adaptive mechanisms are obtained using convolutions of PLDs~\citep[see also][Thm.\;1]{sommer2019privacy}.
\begin{thm} \label{thm:integral}
Consider a $k$-fold non-adaptive composition of a mechanism $\mathcal{M}$. 
The composition is tightly $(\veps,\delta)$-DP for $\delta(\veps)$ given by
$$
\delta(\veps) = \max \{ \delta_{X/Y}(\veps), \delta_{Y/X}(\veps) \},
$$ 
where
\begin{equation*} 
	\begin{aligned}
		\delta_{X/Y}(\veps) & = 1 - \big(1-\delta_{X/Y}(\infty)\big)^k +  \\
	&	\int_\veps^\infty (1 - \ee^{\veps - s})\left(\omega_{X/Y} *^k \omega_{X/Y} \right) (s)  \, \dd s,
	\end{aligned}
\end{equation*}
where $\delta_{X/Y}(\infty)$ is as defined in \eqref{eq:delta_inf} and 
$\omega_{X/Y} *^k \omega_{X/Y}$ denotes the $k$-fold convolution of 
the density function $\omega_{X/Y}$ (an analogous expression holds for $\delta_{Y/X}(\veps)$).
\end{thm}
We remark that our approach also allows computing tight privacy bounds for a composite mechanism $\mathcal{M}_1 \circ \ldots \circ \mathcal{M}_k$, 
where the PLDs of the mechanisms $\mathcal{M}_{i}$ vary (see the supplementary material).

\subsection{Subsampling Amplification} \label{sec:subsampling}

The subsampling amplification can be analysed similarly as by~\cite{koskela2020} in the case of the Gaussian mechanism.
For example, considering the $\sim_R$-neighbouring relation and 
using the Poisson subsampling with subsampling ratio $0<q<1$ leads to considering the pair of density functions
$$
q \cdot f_X  + (1-q) \cdot f_Y \quad \textrm{and} \quad f_Y,
$$ 
where the density function 
$f_X$ corresponds to a subsample including the additional data element. 
Subsampling without and with replacement using $\sim_S$-neighbouring relation can be 
analysed with mixture distributions analogously~\citep{koskela2020}.

\section{Fourier Accountant for Discrete-Valued Mechanisms}

We next describe the numerical method for computing tight DP guarantees for discrete one-dimensional
distributions using the PLD formalism.  
We will apply the fast Fourier transform to numerically evaluate the PLD convolutions of Theorem~\ref{thm:integral}.

\subsection{Fast Fourier Transform}

Let
$$
x = \begin{bmatrix} x_0,\ldots,x_{n-1} \end{bmatrix}^\mathrm{T}, \, w =  \begin{bmatrix} w_0,\ldots,w_{n-1} \end{bmatrix}^\mathrm{T} \in \mathbb{R}^n.
$$
The discrete Fourier transform $\mathcal{F}$ and its inverse $\mathcal{F}^{-1}$ are defined as~\citep{stoer_book}
\begin{equation*} 
	\begin{aligned}
	(\mathcal{F} x)_k  &= \sum\nolimits_{j=0}^{n-1} x_j \ee^{- \mathrm{i} \, 2 \pi k j / n},  \\
	(\mathcal{F}^{-1} w  )_k  &= \frac{1}{n} \sum\nolimits_{j=0}^{n-1} w_j \ee^{ \mathrm{i} \, 2 \pi k j / n},
	\end{aligned}
\end{equation*}
where $\mathrm{i} = \sqrt{-1}$.
Evaluating $\mathcal{F} x$ and $\mathcal{F}^{-1}w$  naively takes $O(n^2)$ operations,
however evaluation using the Fast Fourier Transform (FFT)~\citep{cooley1965} reduces
the running time complexity to $O(n \log n)$. 

For our purposes FFT will be useful as it enables evaluating the discrete convolutions 
efficiently. The so-called convolution theorem~\citep{stockham1966} states that for periodic discrete convolutions 
it holds that 
\begin{equation} \label{eq:conv_thm}
	\sum\nolimits_{i=0}^{n-1} v_i w_{k-i} = \mathcal{F}^{-1} ( \mathcal{F} v \odot \mathcal{F} w),
\end{equation}
where $\odot$ denotes the elementwise product and the summation indices are  modulo $n$. 
Using \eqref{eq:conv_thm}, repeated convolutions are evaluated efficiently.

\subsection{Grid Approximation}

In order to harness the FFT, we place the PLD on a grid
\begin{equation} \label{eq:grid}
X_n = \{x_0,\ldots,x_{n-1}\}, \quad n \in \mathbb{Z}^+,
\end{equation}
where
$$
	x_i = -L + i \Delta x, \quad  \Delta x = 2L/n. 
$$
Suppose the distribution $\omega$ of the PLD is of the form 
$$
\omega(s) = \sum\nolimits_{i=0}^{n-1} a_i \cdot \deltas{s_i},
$$
where $a_i \geq 0$ and $-L \leq s_i \leq L - \Delta x$, $0 \leq i \leq n-1$. 
We define the grid approximations
\begin{equation} \label{eq:omegaRL}
	\begin{aligned}
		\omega^\mathrm{L}(s) & := \sum\nolimits_{i=0}^{n-1} a_i \cdot \deltas{s_i^\mathrm{L}},   \\
		\omega^\mathrm{R}(s) & := \sum\nolimits_{i=0}^{n-1} a_i \cdot \deltas{s_i^\mathrm{R}}, 
	\end{aligned}
\end{equation}
where
\begin{equation*}
	\begin{aligned}
		&   s_i^\mathrm{L} = \max \{  x \in X_n \, : \, x \leq s_i  \}, \\
	    &   s_i^\mathrm{R} = \min \{  x \in X_n \, : \, x \geq  s_i\},
	\end{aligned}
\end{equation*}
i.e., $s_i^L$ and $s_i^R$ refer to the closest left and right grid approximation points to $s_i$.
We note that as $s_i$'s correspond to the log ratios of probabilities of individual events,
often a moderate $L$ is sufficient for the condition $-L \leq s_i \leq L - \Delta x$ to hold for all $i$.
In the Supplements we provide analysis also for the case where this assumption does not hold.
From \eqref{eq:omegaRL} we have: 
\begin{lem} \label{lem:deltaineq}
Let $\delta(\veps)$ be given by the integral formula of Lemma~\ref{lem:maxrepr1} 
and let $\delta^\mathrm{L}(\veps)$ and $\delta^\mathrm{R}(\veps)$ be determined analogously by $\omega^\mathrm{L}$ and $\omega^\mathrm{R}$.
Then for all $\veps>0$ :
\begin{equation*} 
	\delta^\mathrm{L}(\veps) \leq \delta(\veps) \leq \delta^\mathrm{R}(\veps).
\end{equation*}
\end{lem}
Lemma~\ref{lem:deltaineq} directly generalises to convolutions. 
The following bounds for the moment generating functions will be used in the error analysis.

\begin{lem} \label{lem:mgfineq}
Let $\omega, \, \omega^\mathrm{R}$ and $\omega^\mathrm{L}$ also denote the random variables determined by the 
density functions defined above, and let $0 < \lambda < (\Delta x)^{-1}$. Then
\begin{equation*}
	\mathbb{E} [ \ee^{ \lambda \omega^\mathrm{L}} ] \leq \mathbb{E} [ \ee^{ \lambda \omega} ], \quad 
	\mathbb{E} [ \ee^{  - \lambda \omega^\mathrm{L}} ] \leq \tfrac{1}{1-\lambda \Delta x} \mathbb{E} [ \ee^{ - \lambda \omega} ]
\end{equation*}	
and
\begin{equation*} 
	\mathbb{E} [ \ee^{ - \lambda \omega^\mathrm{R}} ] \leq \mathbb{E} [ \ee^{ - \lambda \omega } ], \quad
	\mathbb{E} [ \ee^{ \lambda \omega^\mathrm{R}} ] \leq \tfrac{1}{1-\lambda \Delta x} \mathbb{E} [ \ee^{ \lambda \omega } ].
\end{equation*}
\end{lem}

\subsection{Truncation of Convolutions and Periodisation}

The FFT assumes that inputs are periodic over a finite range. 
We describe truncation of convolutions and periodisation of distribution functions to meet this assumption.
Suppose $\omega$ is defined such that 
\begin{equation} \label{eq:omega}
	\omega(s) = \sum\nolimits_i a_i \cdot \deltas{s_i},
\end{equation}
where $a_i \geq 0$ and $s_i = i \Delta x$. 
The convolutions can then be written as 
\begin{equation*}
	\begin{aligned}
	(\omega * \omega )(s)  &=  \sum\nolimits_{  i,j} a_i a_j \cdot \deltas{s_i + s_j}  \\
	 &= \sum\nolimits_i \Big(\sum\nolimits_j a_j a_{i-j} \Big) \cdot \deltas{s_i}.
\end{aligned}
\end{equation*}
Let $L>0$. We truncate these convolutions to the interval $[-L,L]$ such that 
\begin{equation*}
	\begin{aligned}
	(\omega * \omega )(s) & \approx \sum\nolimits_i \Big(\sum\nolimits_{-L \leq s_j < L} a_j a_{i-j} \Big) \cdot \deltas{s_i}   \\
	& =: (\omega \circledast \omega )(s).
\end{aligned}
\end{equation*}
We define $\widetilde{\omega}$ to be a $2 L$-periodic extension of $\omega$, i.e., $\widetilde{\omega}$ is of the form
$$
\widetilde{\omega}(s) = \sum\nolimits_{m \in \mathbb{Z}} \, \sum\nolimits_i a_i \cdot \deltas{s_i + m \cdot 2 L}.
$$
We further approximate 
$$
(\omega \circledast \omega )  \approx (\widetilde{\omega} \circledast \widetilde{\omega} ).
$$
In case the distribution $\omega$ is defined on an equidistant grid, FFT can be used to evaluate $\widetilde{\omega} \circledast \widetilde{\omega} $
as follows:
\begin{lem} \label{lem:fft}
Let $\omega$ be of the form \eqref{eq:omega}, such that
$n$ is even, $L>0$, $\Delta x = 2L/n$ and $s_i = -L + i \Delta x$, $0 \leq i \leq n-1$. 
Define
$$
\boldsymbol{a} = \begin{bmatrix} a_0 & \ldots & a_{n-1} \end{bmatrix}^\mathrm{T} \quad \textrm{and} \quad
 D = \begin{bsmallmatrix} 0 & I_{n/2} \\ I_{n/2} & 0 \end{bsmallmatrix} \in \mathbb{R}^{n \times n}.
$$
Then, 
$$
(\widetilde{\omega} \circledast^k \widetilde{\omega} )(s) = \sum\nolimits_{i=0}^{n-1} b_i^k \cdot \deltas{s_i},
$$
where
$$
b_i^k = \left[D \, \mathcal{F}^{-1} \big(\mathcal{F}( D \boldsymbol{a} )^{ \odot k}   \big) \right]_i,
$$
and $^{ \odot k}$ denotes the elementwise power of vectors.
\end{lem}

\subsection{Approximation of the $\delta(\veps)$-Integral} \label{subsec:desc}

Finally, using the truncated and periodised convolutions we approximate the integral formula in Lemma~\ref{lem:maxrepr1}
for the tight $\delta$-value as
\begin{equation} \label{eq:approx_int}
	\begin{aligned}
	    & \int_\veps^\infty (1 - \ee^{\veps - s})(\omega *^k \omega ) (s)  \, \dd s \\
	   \approx    & \int_\veps^L (1 - \ee^{\veps - s})(\widetilde{\omega} \circledast^k \widetilde{\omega} ) (s)  \, \dd s \\
	   = & \sum\nolimits_{\ell=\ell_\veps}^{n-1}  \big(1 - \ee^{\veps - ( - L + \ell \Delta x)} \big) \, b^k_\ell,
	\end{aligned}
\end{equation}
where $\ell_\veps = \min \{ \ell \in \mathbb{Z} \, : \, -L + \ell \Delta x > \veps \}$ and the vector $b^k \in \mathbb{R}^n$
is given by Lemma~\ref{lem:fft}.
We describe the method in the pseudocode of Algorithm~\ref{alg:delta}. 
In the following section we give
an error bound for the approximation with respect to the parameter $L$. 

\begin{algorithm}[ht!]
\caption{ Fourier Accountant Algorithm for Discrete-Valued Mechanisms}
\begin{algorithmic}
\STATE{Input: distribution $\omega$ of the form \eqref{eq:omega}, such that
$n$ is even and $s_i = -L + i \Delta x$, $0 \leq i \leq n-1$, $\Delta x=2L/n$,
 number of compositions $k$.}
\vspace{2mm}
\STATE{Set
$$
\boldsymbol{a} = \begin{bmatrix} a_0 & \ldots & a_{n-1} \end{bmatrix}^\mathrm{T}, \quad D = \begin{bsmallmatrix} 0 & I_{n/2} \\ I_{n/2} & 0 \end{bsmallmatrix}.
$$
}
\STATE{Evaluate the convolutions using Lemma~\ref{lem:fft} and FFT:
\begin{equation*}
\boldsymbol{b}^k = \left[D \, \mathcal{F}^{-1} \big(\mathcal{F}( D \boldsymbol{a} )^{\odot k}   \big) \right],
\end{equation*}
} \STATE{Determine the starting point of the integral interval:
$$
\ell_\veps = \min \{ \ell \in \mathbb{N} \, : \, -L + \ell \Delta x > \veps \},
$$
} \STATE{Approximate $\delta(\veps)$ using Lemma~\ref{lem:maxrepr1}:
\begin{equation*}
	\begin{aligned}
	&	\delta( \veps) \approx  1-(1-\delta_{X/Y}(\infty))^k \\
	&+  \sum\nolimits_{\ell=\ell_\veps}^{n-1}  \big(1 - \ee^{\veps - ( - L + \ell \Delta x)} \big) \, b^k_\ell.	
	\end{aligned}
\end{equation*}
}
\end{algorithmic}
\label{alg:delta}
\end{algorithm}

%
\begin{remark}
To evaluate $\veps$ as a function of $\delta$,
Newton's method can be used~\citep{koskela2020}. Suppose $\omega$ is continuous and $\delta(\veps)$ given by the integral~\eqref{eq:approx_int}. 
Then,
$\delta'(\veps) = - \int_\veps^\infty \ee^{\veps - s}(\omega *^k \omega ) (s)  \, \dd s$
and Newton's method applied to the function $\delta(\veps) - \bar{\delta}$ gives the iteration
\begin{equation} \label{eq:newton}
	\veps_{\ell+1} = \veps_\ell -  \frac{\delta(\veps_\ell) - \bar{\delta}} {\delta'(\veps_\ell) }.
\end{equation}
Similarly to~\eqref{eq:approx_int} this naturally translates to the case of discrete distributions.
We use as a stopping criterion $\abs{\delta({\veps_{\ell}}) - \bar{\delta}} \leq \tau	$
for some prescribed tolerance parameter $\tau$ and an initial value $\veps_0=0$.
In experiments, for an equal stopping criterion $\tau$, the iteration \eqref{eq:newton} 
gave more than twice as fast convergence as the binary search algorithm.
\end{remark}

\section{Error Analysis} \label{sec:err_est}

We next give a bound for the error induced by Algorithm~\ref{alg:delta} which is determined by the parameter $L$. 
The total error consists of (see the supplementary material)
\begin{enumerate}
	\item The tail integral $\int_L^\infty (\omega *^k \omega ) (s)  \, \dd s$.
	\item The error arising from periodisation of $\omega$ and truncation of the convolutions. 
\end{enumerate}
We obtain bounds for these two error sources using the Chernoff bound~\citep{wainwright2019}
$$\mathbb{P}[ X \geq t] 
\leq \frac{ \mathbb{E}[ \ee^{\lambda X} ] }{\ee^{\lambda t}}$$
which holds for any random variable $X$ and all $\lambda > 0$. Suppose $\omega_{X/Y}$ is of the form 
\begin{equation} \label{eq:omega_xy}
	\omega_{X/Y}(s) = \sum\nolimits_{i=0}^{n-1} a_{X,i} \cdot \deltas{s_i},  
\end{equation} 
where $s_i = \log \left( \tfrac{a_{X,i}}{a_{Y,i}} \right)$ and $a_{X,i},a_{Y,i}>0$.  
Then, the moment generating function of $\omega_{X/Y}$ is given by
\begin{equation} \label{eq:pld_lmf}
	\begin{aligned}
		\mathbb{E} [\ee^{\lambda \omega_{X/Y} }] &= \int_{-\infty}^\infty \ee^{\lambda s} \omega(s) \, \dd s \\
		& = \sum\nolimits_{i=0}^{n-1}  \ee^{\lambda s_i} \cdot a_{X,i} \\
		& = \sum\nolimits_{i=0}^{n-1}  \left( \frac{a_{X,i}}{a_{Y,i}}  \right)^\lambda a_{X,i}.
	\end{aligned}
\end{equation}

\subsection{Connection to RDP} \label{subsubsec:renyi}

Suppose $f_X(t) = \sum_i a_{X,i} \cdot \deltat{t_i}$,
$f_Y(t) = \sum_i a_{Y,i} \cdot \deltat{t_i}$
for some coefficients $a_{X,i},a_{Y,i}$,
and suppose $\omega_{X/Y}$ is of the form \eqref{eq:omega_xy}.
Then, we have that 
$$
\mathbb{E} [\ee^{\lambda \omega_{X/Y} }] = \lambda \cdot D_{\lambda+1}(f_X,f_Y),
$$
where $D_\alpha$ denotes the R\'enyi divergence of order $\alpha$~\citep{mironov2017}.
Further, defining 
$$
\alpha(\lambda) := \log (\mathbb{E} [\ee^{\lambda \omega_{X/Y} }]),
$$ 
we see that $\alpha(\lambda)$
is exactly the logarithm of the moment generating function of the privacy loss function
as defined, e.g., by~\citet{Abadi2016} and~\citet{mironov2019}. Thus
existing R\'enyi differential privacy estimates for $\alpha(\lambda)$ could be used to bound
the moment generating function of $\omega_{X/Y}$.


\subsection{Tail Bound} \label{subsubsec:num_approx1}

Denote $S_k := \sum_{i=1}^k \omega_i$, where $\omega_i$ denotes the PLD random variable of the $i$th mechanism.
If $\omega_i$'s are independent, we have that 
$$ 
\mathbb{E} [ \ee^{\lambda S_k}  ]  = \prod\nolimits_{i=1}^k \mathbb{E} [ \ee^{\lambda \omega_i}  ].
$$
Then, if $\omega_i$'s are i.i.d. and distributed as $\omega$, the Chernoff bound shows that for any $\lambda > 0$
\begin{equation} \label{eq:tail_bound}
	\begin{aligned}
		\int_L^\infty  ( \omega \ast^k \omega)(s) \, \dd s &= \mathbb{P}[ S_k \geq L ] \\ &\leq 
		\prod\nolimits_{i=1}^k \mathbb{E} [ \ee^{\lambda \omega_i}  ] \,  \ee^{- \lambda L} \\
		& \leq \ee^{k \alpha(\lambda)}  \ee^{- \lambda L},
	\end{aligned}
\end{equation}
where $\alpha(\lambda) = \log( \mathbb{E} [\ee^{\lambda \omega }] )$.

\subsection{Total Error}

We define $\alpha^+(\lambda)$ and $\alpha^-(\lambda)$ via the moment generating function 
of the PLD as
$$
\alpha^+(\lambda) = \log \Big(	\mathbb{E} [\ee^{\lambda \omega }] \Big),
\quad  \alpha^-(\lambda) = \log \Big(	\mathbb{E} [\ee^{ - \lambda \omega }] \Big).
$$
Using the analysis given in the supplementary material, we bound the errors arising from the
periodisation of the distribution and truncation of the convolutions. As a result, combining with \eqref{eq:tail_bound}, we obtain
the following bound for the total error incurred by Algorithm~\ref{alg:delta}.
\begin{thm} \label{thm:alg_error_bound}
Let $\omega$ be defined on the grid $X_n$ as described above, let $\delta(\veps)$ give the tight $(\veps,\delta)$-bound for $\omega$
and let $\widetilde{\delta}(\veps)$ be the result of Algorithm~\ref{alg:delta}.
Then, for all $\lambda > 0$
\begin{equation*}
	\begin{aligned}
 \abs{ \delta(\veps)  -  \widetilde{\delta}(\veps) }
 \leq	& \bigg(  \frac{2 \ee^{(k+1) \alpha^+(\lambda)} -\ee^{k \alpha^+(\lambda)}  -   \ee^{ \alpha^+(\lambda)}}{\ee^{\alpha^+(\lambda)} - 1} \\
		+&  \frac{ \ee^{(k+1) \alpha^-(\lambda)} - \ee^{ \alpha^-(\lambda)}}{\ee^{\alpha^-(\lambda)}-1} \bigg) \, \frac{\ee^{-L \lambda}}{1- \ee^{-L \lambda}}.
 	\end{aligned}
\end{equation*}	
\end{thm}


Given a discrete-valued PLD distribution $\omega$, we get strict lower and upper $\delta(\veps)$-DP bounds as follows.
Using parameter values $L>0$ and $n \in \mathbb{Z}^+$, we form a grid $X_n$ as defined in \eqref{eq:grid} and 
place $\omega$ on $X_n$ to obtain $\omega^\mathrm{L}$ and $\omega^\mathrm{R}$ as defined in \eqref{eq:omegaRL}. 
We then approximate $\delta^L(\veps)$ and $\delta^R(\veps)$
using Algorithm~\ref{alg:delta}. We estimate the error incurred by the approximation using 
Thm.~\ref{thm:alg_error_bound} and the expressions given by Lemma~\ref{lem:mgfineq}.
By subtracting this error from the approximation of $\delta^L(\veps)$ and adding it to the approximation of $\delta^R(\veps)$ 
and using Lemma~\ref{lem:deltaineq}, we obtain strict lower and upper bounds for $\delta(\veps)$.

To obtain $\alpha^+(\lambda)$ and $\alpha^-(\lambda)$, 
we evaluate  the moment generating functions $\mathbb{E} [\ee^{\lambda \omega }]$ and $\mathbb{E} [\ee^{ - \lambda \omega }]$
using the finite sum \eqref{eq:pld_lmf}. We use $\lambda=L/2$ in all experiments.

We emphasise that the error analysis is given in terms of the parameter $L$.
The parameter $n$ can be increased in case the resulting lower and upper bounds for $\delta(\veps)$ are too far from each other.

%

%
%

\section{Examples}

\subsection{The Exponential Mechanism}

\begin{figure} [h!]
	\begin{center}
  \includegraphics[width=0.7\linewidth]{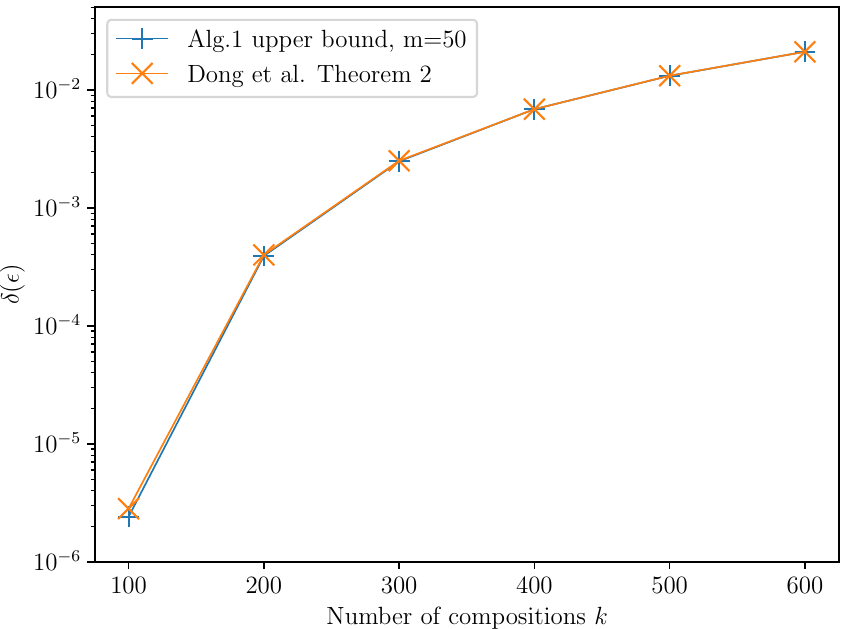}
	\caption{Exponential mechanism with a counting query quality score and parameter value $\veps=1.0$.
	We compute $\delta(\veps)$ using Algorithm~\ref{alg:delta} and the optimal bound given by~\citet[Thm.\;2]{dong2020}, for $\widetilde{\veps}=0.1$.}
\label{fig:1}
	\end{center}
\end{figure}

\begin{figure} [h!]
	\begin{center}
  \includegraphics[width=0.7\linewidth]{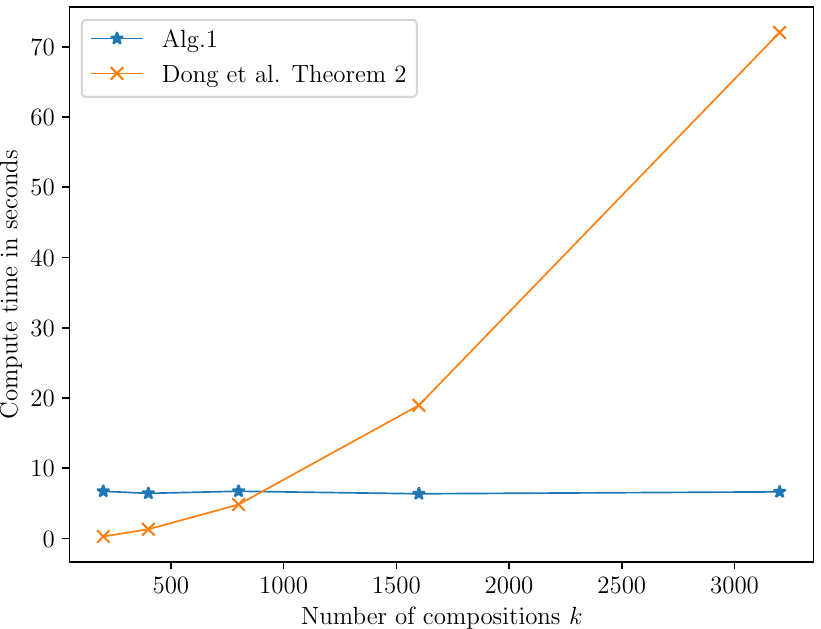}
	\caption{Compute times for different number of compositions $k$, using Algorithm~\ref{alg:delta} and the expression of~\citet[Thm.\;2]{dong2020} $\delta(\veps)$.
	The evaluation of the expression by~\citet[Thm.\;2]{dong2020} is optimised using the logarithmic gamma function. }
\label{fig:2}
	\end{center}
\end{figure}

Consider the exponential mechanism $\mathcal{M}$ with quality score 
$u \, : \, \mathcal{X}^n \times \mathcal{Y} \rightarrow \mathbb{R}$ and parameter $\widetilde{\veps}$, i.e., 
an outcome $y$ is sampled with probability
$$
\mathbb{P}(\mathcal{M}(X)=y) = \frac{ \ee^{\widetilde{\veps} u(X,y)} }{ \sum_y \ee^{\widetilde{\veps} u(X,y)} }.
$$
Consider the neighbouring relation $\sim_R$. 
Let $u$ be a counting query, i.e., 
$$
u(X,y) = \sum\nolimits_{x \in X} \mathbf{1}({x = y}),
$$
and let $\mathcal{Y} = \{0,1\}$.
Denote by $m$ the number of elements in $X$ which equal $0$. 
Let $Y \in  \mathcal{X}^{n-1}$, $X \sim Y$, be such that $m-1$ elements equal $0$.
Then, the logarithmic ratio at $y=0$ is given by
\begin{equation*}
	\begin{aligned}
	s_0	&:= \log \Bigg( \frac{ \mathbb{P}(\mathcal{M}(X) = 0)}{ \mathbb{P}(\mathcal{M}(Y) = 0) } \Bigg) \\
	 & = \log \Bigg( \frac{\ee^{\widetilde{\veps} m}}{ \ee^{\widetilde{\veps} (m-1)} } 
	\frac{ \ee^{\widetilde{\veps} (m-1)} + \ee^{\widetilde{\veps} (n-m)}  }{ \ee^{\widetilde{\veps} m} + \ee^{\widetilde{\veps} (n-m)} } \Bigg) 
	\end{aligned}
\end{equation*}
and similarly 
$
s_1 = \log \left( \tfrac{ \mathbb{P}(\mathcal{M}(X) = 1) }{  \mathbb{P}(\mathcal{M}(Y) = 1) } \right).
$
Using the values of $\mathbb{P}(\mathcal{M}(X) = i)$
and $s_i$, $i=0,1$, we obtain the PLD. 
We set $\widetilde{\veps}=0.05$ and $m=50$.
Figure~\ref{fig:1} shows the $\delta(\veps)$-values for $\veps=1.0$, when computed using
Algorithm~\ref{alg:delta} for $\mathcal{M}(X)$ and $\mathcal{M}(Y)$ and
the optimal bound~\cite[Thm.\;2]{dong2020}. The corresponding compute times are shown in Figure~\ref{fig:2}.
The evaluation of the expression in~\cite[Thm.\;2]{dong2020} is optimised using the logarithmic gamma function. 


\subsection{The Binomial Mechanism}

The binomial mechanism by~\citet{agarwal2018} adds binomially distributed 
noise $Z$ with parameters $n \in \mathbb{N}$ and $0<p<1$ to the output of a query $f$ with output space $\mathbb{Z}^d$ 
as
$$
\mathcal{M}(X) = f(X) + (Z - np) \cdot s,
$$
where $s=1/j$ for some $j\in \mathbb{N}$ and where for each coordinate $i$, $Z_i \sim \mathrm{Bin}(n,p)$ and 
$Z_i$'s are independent.

As described in the proof of Thm. 1 of~\citet{agarwal2018}, for the privacy analysis of the binomial mechanism it is sufficient
to consider the neighbouring binomial distributions centred at 0 and $\Delta$.
If, for example, $d=1$, it is sufficient to consider the neighbouring binomial distributions
\begin{equation*}
	\begin{aligned}
		f_X(t) &= \sum\nolimits_{i=0}^n \binom{n}{i} p^i (1-p)^{n-i} \deltat{i+\Delta}, \\
		f_Y(t) &= \sum\nolimits_{i=0}^n \binom{n}{i} p^i (1-p)^{n-i} \deltat{i}.
	\end{aligned}
\end{equation*}
Then, the privacy loss distribution $\omega_{X/Y}$ is of the form
\begin{equation*}
	\begin{aligned}
\omega_{X/Y}(s) &= \sum\nolimits_{i=\Delta}^{n-\Delta} a_i \cdot \deltas{s_i},  \\
		a_i &= \binom{n}{i} p^{i-\Delta} (1-p)^{n-i+\Delta}, \\
		s_i &= \log \left( \tfrac{ \binom{n}{i}  }{ \binom{n}{i-\Delta} } \left(\tfrac{1-p}{p}\right)^\Delta \right).
	\end{aligned}
\end{equation*}
Moreover,
$$
\omega_{X/Y}(\infty) = \sum\nolimits_{i=n-\Delta+1}^n  \binom{n}{i} p^i (1-p)^{n-i},
$$
and determining the privacy loss distribution $\omega_{Y/X}$ can be done analogously.

The $(\veps,\delta)$-analysis of the multivariate binomial mechanism can be carried out 
via one-dimensional distributions using the following observation. 
\begin{thm} \label{thm:equivalence}
Consider a function $f \, : \, \mathcal{X}^N \rightarrow \mathbb{R}^d$ and a randomised mechanism
$\mathcal{M}$ of the form
$
\mathcal{M}(X) = f(X) + Z,
$
where $Z_i$'s are independent random variables. Suppose the data sets $X$ and $Y$ lead to the $\delta(\veps)$-upper bound, and denote $\Delta = f(X) - f(Y)$.
Then, the tight $(\veps,\delta)$-bound for $\mathcal{M}$ 
is given by the tight $(\veps,\delta)$-bound for the non-adaptive compositions of one-dimensional random variables
$$
\Delta_i + Z_i \quad \textrm{and} \quad Z_i, \quad 1 \leq i \leq d.
$$
\end{thm}

\begin{figure} [h!]
	\begin{center}
  \includegraphics[width=0.7\linewidth]{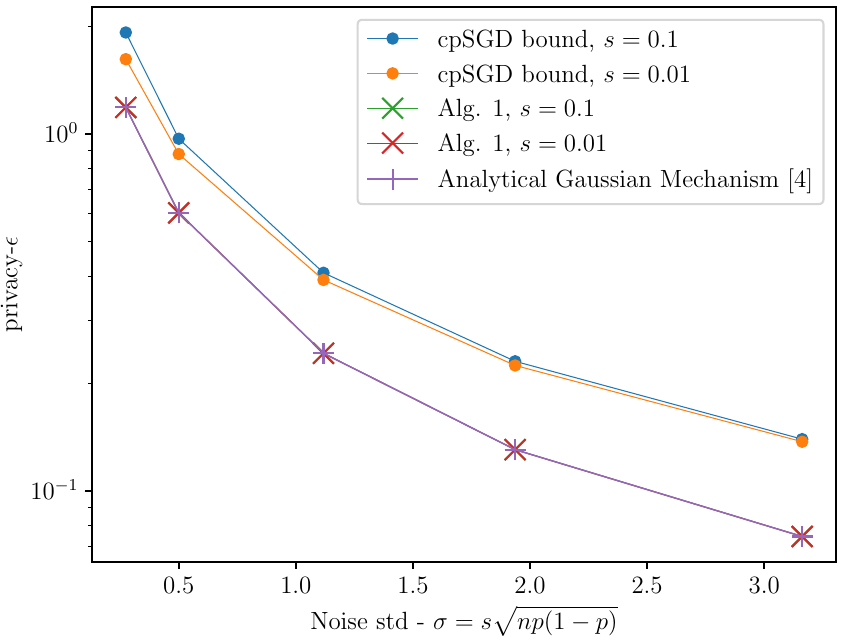}
	\caption{Comparison of the cpSGD bound~\cite[Thm.\;1]{agarwal2018} and the upper bound given by Algorithm~\ref{alg:delta}
 	  ($\delta=10^{-4}$, $p=0.5$). The bound given by Algorithm~\ref{alg:delta} is close to that of the Analytical Gaussian mechanism~\citep{balle2018gauss}.}
\label{fig:binom1}
	\end{center}
\end{figure}

Figure~\ref{fig:binom1} illustrates how Algorithm~\ref{alg:delta} gives tighter bounds than the bound of~\citet[Thm.\;1]{agarwal2018},
and also how the $(\veps,\delta)$-bound given by Algorithm~\ref{alg:delta} is close to the tight bound of the Gaussian mechanism
for the corresponding variance~\citep[Analytical Gaussian mechanism by][]{balle2018gauss}.
We use an example analogous to~\citet[Fig. 1]{agarwal2018}:
we set $\Delta= \begin{bmatrix} \tfrac{1}{10}, \ldots, \tfrac{1}{10} \end{bmatrix}^\mathrm{T} \in \mathbb{R}^{100}$, $p=0.5$ and vary the parameters $n$ and $s$.
Using Thm.~\ref{thm:equivalence}, we obtain tight $(\veps,\delta)$-bounds by considering a 100-fold compositions of one-dimensional mechanisms
$$ 
\mathcal{M}(X) = \tfrac{1}{10} + (Z - np) \cdot s, \quad \mathcal{M}(Y) = (Z - np),
$$
and thus we can use Algorithm~\ref{alg:delta} to obtain tight $(\veps,\delta)$-bounds for a single call of $\mathcal{M}(X)$.

\begin{figure} [h!]
	\begin{center}
  \includegraphics[width=0.7\linewidth]{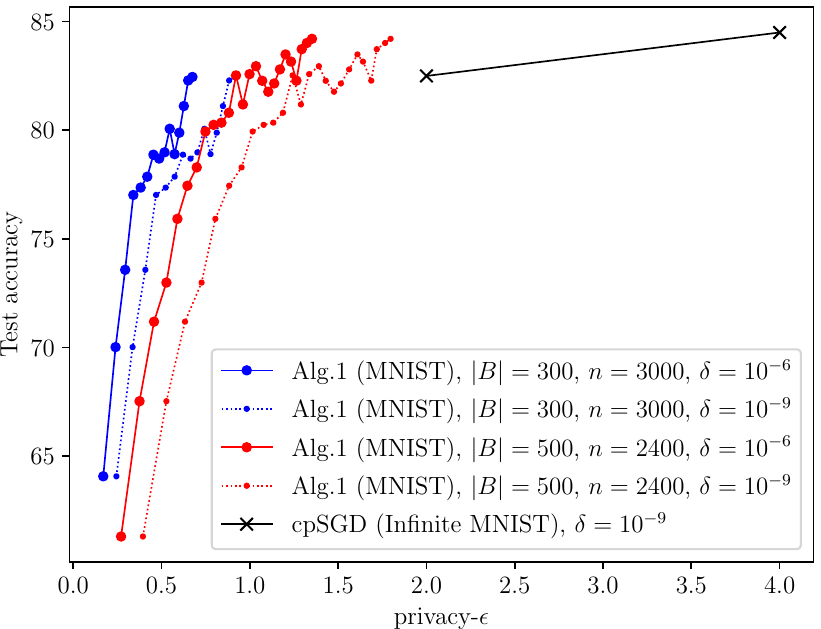}
	\caption{A small feedforward model run on MNIST ($M=6 \cdot 10^4$) using Algorithm 1
		and on Infinite MNIST ($M=2.5 \cdot 10^8$) using cpSGD~\citep{agarwal2018}.
		Algorithm~\ref{alg:delta} takes into account the subsampling amplification (Sec.~\ref{sec:subsampling}).}
\label{fig:binom2}
	\end{center}
\end{figure}

Figure~\ref{fig:binom2} shows results for an MNIST classification task, where we use 
a three-layer feedforward network with ReLUs and a hidden layer of width 60. 
DP-SGD approximation of the gradients is carried out
such that for each per example gradient we use a sign approximation:
the 200 largest elements (by magnitude) of the input layer are approximated by their sign and the rest are set to zero and
similarly the 20 largest of the hidden layer and the largest one of the output layer. Elementwise zero centred binomial noise
with parameters $n$ and $p=0.5$ is then added to the averaged gradients.
By Thm.~\ref{thm:equivalence} and subsampling amplification (Sec.~\ref{sec:subsampling}), 
the $(\veps,\delta)$-bound can be obtained by running Algorithm~\ref{alg:delta}
for the PLD determined by the distributions
$$
q \cdot f_X + (1-q) \cdot f_Y, \quad \textrm{and} \quad f_Y, 
$$
where $f_X$ and $f_Y$ are the density functions of the random variables
$$ 
X \sim  \mathbf{1} + (Z - np) \quad \textrm{and} \quad Y \sim (Z - np),
$$
where $\mathbf{1} = \begin{bmatrix} 1, \ldots, 1 \end{bmatrix}^\mathrm{T} \in \mathbb{R}^{221}$ and 
for each $i$, $Z_i \sim \mathrm{Bin}(n,p)$ and $Z_i$'s are independent.
Here $q$ denotes the subsampling ratio,
i.e., $q=\abs{B}/M$, where $\abs{B}$ is the minibatch size and $M$ the total size of the training data.
We obtain tight $(\veps,\delta)$-bounds for the training of the network as follows (details in the Supplements). 
We obtain the PLD $\omega$ determined by the distributions
$q \cdot f_X + (1-q) \cdot f_Y$ and $f_Y$ from the PLD determined by $f_X$ and $f_Y$ (that is obtained using Thm.~\ref{thm:equivalence} and Alg.~\ref{alg:delta},
as in the example of Figure~\ref{fig:binom1}). We then apply Algorithm~\ref{alg:delta} to $\omega$, for a given number of compositions.

The results of Figure~\ref{fig:binom2} are averages of 5 runs.
We set the initial learning rate $\eta=0.02$.
We linearly decrease the learning rate $\eta$ after each epoch such that it is zero at the end of the training
(when $\abs{B}=500$ starting from epoch 13, and when $\abs{B}=300$ starting from epoch 5).
We compare this method to cpSGD~\citep{agarwal2018} applied to Infinite MNIST data set which has the same test data set as MNIST.
The results for cpSGD are extracted from~\citet[Fig. 2]{agarwal2018}. For $\veps=2.0$ we extract the result where each element of the gradient
requires 8 bits and for $\veps=4.0$ the one requiring 16 bits. We note that when $n=3000$ our method requires 
12 bits per element. 

\subsection{The Subsampled Gaussian Mechanism} \label{subsec:subsampled}

We next show how to compute rigorous DP bounds for the subsampled Gaussian mechanism 
using the method presented here. We consider the Poisson subsampling and $\sim_R$-neighbouring relation.
For a subsampling ratio $q$ and noise level $\sigma$, the continuous PLD is given by~\cite{koskela2020}
\begin{equation} \label{eq:subsampled_PLD_def} 
\omega(s) = \begin{cases}
f(g(s))g'(s), &\text{ if }  s > \log(1-q), \\
0, &\text{ otherwise},
\end{cases}
\end{equation}
where
$$
f(t) = \frac{1}{\sqrt{2 \pi \sigma^2}} \, [ q \ee^{ \frac{-(t-1)^2}{2 \sigma^2}} + (1-q) \ee^{-\frac{t^2}{2 \sigma^2}} ]
$$
and
$$
g(s) = \sigma^2 \log \left( \frac{\ee^s - (1-q)}{q} \right) + \frac{1}{2}.
$$
Let $L>0$, $n \in \mathbb{Z}^+$, $\Delta x = 2L/n$ and $s_i = -L + i \Delta x$ for all $i \in \mathbb{Z}$. 
We define
\begin{equation} \label{eq:c_plusminus1}
	\begin{aligned}
		\omega_{\mathrm{min}}(s) &= \sum\nolimits_{i=0}^{n-1} c^-_i \cdot \deltas{s_i},  \\
		\omega_{\mathrm{max}}(s) &= \sum\nolimits_{i=0}^{n-1} c^+_i \cdot \deltas{s_i},
	\end{aligned}
\end{equation}
where
\begin{equation*} 
	\begin{aligned}
		c^-_i &= \Delta x \cdot \min\nolimits_{s \in [s_i, s_{i+1}]} \omega(s), \\
		c^+_i &= \Delta x \cdot \max\nolimits_{s \in [s_{i-1}, s_i]} \omega(s).
	\end{aligned}
\end{equation*}
Furthermore, we define
\begin{equation} \label{eq:c_plusminus2}
	\begin{aligned}
		\omega^\infty_{\mathrm{min}}(s) &= \sum\limits_{i \in \mathbb{Z}} c^-_i \cdot \deltas{s_i}, \\
		\omega^\infty_{\mathrm{max}}(s) &= \sum\limits_{i \in \mathbb{Z}} c^+_i \cdot \deltas{s_i}.
	\end{aligned}
\end{equation}
We find that $\omega$ as defined in \eqref{eq:subsampled_PLD_def} 
has one stationary point which we determine numerically.
Using this fact, the numerical values of $c^-_i$ and $c^+_i$ can be straightforwardly computed.

We obtain approximations for the lower and upper bounds $\delta_{\mathrm{min}}(\veps)$ and $\delta_{\mathrm{max}}(\veps)$ 
by running Algorithm~\ref{alg:delta} for $\omega^\infty_{\mathrm{min}}$ and $\omega^\infty_{\mathrm{max}}$ using some prescribed parameter values $n$ and $L$:
\begin{lem} \label{lem:subsampled_ineq}
Let $\delta(\veps)$ be given by the integral formula of Thm.~\ref{thm:integral} for some privacy loss distribution $\omega$. 
Let $\delta_{\mathrm{min}}^\infty(\veps)$ and $\delta_{\mathrm{max}}^\infty(\veps)$ be defined analogously by 
$\omega^\infty_{\mathrm{min}}$ and $\omega^\infty_{\mathrm{max}}$.
Then for all $\veps>0$ we have
$$
\delta_{\mathrm{min}}^\infty(\veps) \leq \delta(\veps) \leq \delta_{\mathrm{max}}^\infty(\veps).
$$
\begin{proof}
Supplements.
\end{proof}
\end{lem}
Running Alg.~\ref{alg:delta} for $\omega^\infty_{\mathrm{min}}$ and $\omega^\infty_{\mathrm{max}}$
is equivalent to running it for the truncated distributions $\omega_{\mathrm{min}}$ and $\omega_{\mathrm{max}}$. However, to obtain the bounds of Thm.~\ref{thm:integral}
(and subsequently strict bounds for $\delta(\veps)$), the analysis has to be carried out for
$\omega^\infty_{\mathrm{min}}$ and $\omega^\infty_{\mathrm{max}}$. To this end, we need bounds for the moment generating functions of $-\omega^\infty_{\mathrm{min}}$, $\omega^\infty_{\mathrm{min}}$
$-\omega^\infty_{\mathrm{max}}$ and $\omega^\infty_{\mathrm{max}}$ (where $-\omega(s) := \sum_i a_i \cdot \deltas{-s_i}$ if $\omega(s) = \sum_i a_i \cdot \deltas{s_i}$).
 We can bound the moment generating function of $\omega^\infty_{\mathrm{max}}$
as follows. We note that $\mathbb{E} [\ee^{\lambda \omega_{\mathrm{max}} }]$ can be evaluated numerically.

\begin{lem} \label{lem:mgfs}
Let $0 < \lambda \leq L$ and assume $\sigma \geq 1$ and $\Delta x \leq c \cdot L$, $0<c<1$.
Let $\omega_{\mathrm{max}}$ and $\omega^\infty_{\mathrm{max}}$ be defined as in~\eqref{eq:c_plusminus1} and~\eqref{eq:c_plusminus2}.
The moment generating function of $\omega^\infty_{\mathrm{max}}$ can be bounded as
$$
\mathbb{E} [\ee^{\lambda \omega^\infty_{\mathrm{max}} }] \leq 
\mathbb{E} [\ee^{\lambda \omega_{\mathrm{max}} }] + \mathrm{err}(\lambda,L,\sigma),
$$
where
\begin{equation*} 
	\begin{aligned}
&		\mathrm{err}(\lambda,L,\sigma) =  \\
& \quad		\ee^{c \lambda L } \frac{2}{\sqrt{ \pi} }  \ee^{ - \frac{ \lambda(2C - \lambda) }{2 \sigma^2} } 
				  \mathrm{erfc} \left( \frac{ (1-c) \sigma^2 L + C - \lambda }{ \sqrt{2} \sigma }   \right)
	\end{aligned}
\end{equation*}
and $C = \sigma^2 \log( \frac{1}{2q}) - \frac{1}{2}$.
\begin{proof}
Supplements.
\end{proof}
\end{lem}	
	
An analogous bound holds for the moment generating functions of $-\omega^\infty_{\mathrm{min}}$, $\omega^\infty_{\mathrm{min}}$ and
$-\omega^\infty_{\mathrm{max}}$ (see the Supplements). In the experiments, the effect of the error term $\mathrm{err}(\lambda,L,\sigma)$ was found to be negligible.

\begin{figure} [h!]
	\begin{center}
  \includegraphics[width=0.7\linewidth]{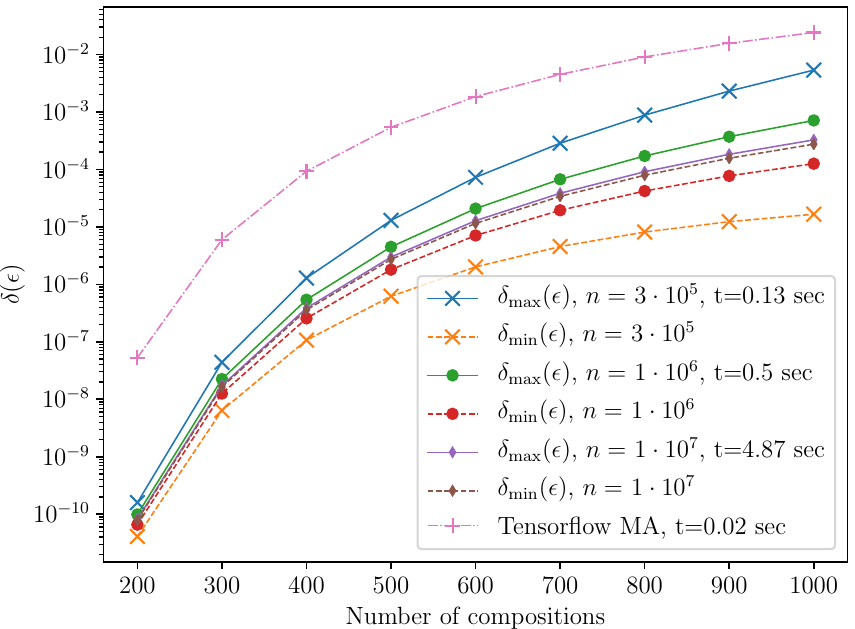}
	\caption{The subsampled Gaussian mechanism and 
	 bounds for $\delta(\veps)$ computed using Algorithm~\ref{alg:delta}, 
	 when $\veps=1.0$, $q=0.02$, $\sigma=2.0$ and $L=8.0$. Here $n$ denotes the number of discretisation points. 
	Compute times are for each curve. }
\label{fig:subsampled}
	\end{center}
\end{figure}

Figure~\ref{fig:subsampled} illustrates the convergence of the bound given by Lemma~\ref{lem:subsampled_ineq}
as $n$ grows and $L$ is fixed.
For comparison, we also show the numerical values given by Tensorflow moments accountant~\citep{Abadi2016}.

\section{Conclusions} 

We have presented a novel approach for computing privacy bounds for discrete-valued mechanisms.
The method provides tools for moments-accountant-like techniques for evaluating
privacy bounds for discrete output DP-SGD algorithms. More specifically, we have shown how to accurately bound the $\delta(\veps)$-DP
for the subsampled binomial mechanism, when the gradients are replaced with a sign approximation.
 Moreover, as the example of Section~\ref{subsec:subsampled}
shows, accurate $(\veps,\delta)$-bounds for continuous mechanisms can also be obtained using the proposed method.
Due to the rigorous error analysis the reported $(\veps,\delta)$-bounds are strict lower and upper privacy bounds.

\section*{Acknowledgements} 

This work has been supported by the Academy of Finland [Finnish Center for Artificial Intelligence FCAI and grants 319264, 325572, 325573].

\bibliography{pld}

\newpage

\appendix

\section{Proofs for the Results of Section 3} 

\subsection{Integral Representation for Exact DP-Guarantees}

Throughout this section we denote for neighbouring datasets
$X$ and $Y$ the density function of $\mathcal{M}(X)$ with $f_X(t)$ and the density function of $\mathcal{M}(Y)$ with $f_Y(t)$.
The definition of approximate differential privacy is equivalently given as follows.
\begin{defn} \label{dfn:tightADP}	
A randomised algorithm $\mathcal{M}$ with an output of one dimensional distributions satisfies 
$(\varepsilon,\delta)$-DP if for every set $S \subset \mathbb{R}$ and every neighbouring datasets $X$ and $Y$
$$
\int\limits_S f_X(t) \, \dd t \leq \ee^\varepsilon \int\limits_S f_Y(t) \, \dd t  + \delta \quad \textrm{and} \quad
\int\limits_S f_Y(t) \, \dd t \leq \ee^\varepsilon \int\limits_S f_X(t) \, \dd t  + \delta.
$$
We call $\mathcal{M}$ tightly $(\varepsilon,\delta)$-DP, if there does not exist $\delta' < \delta$
such that $\mathcal{M}$ is $(\varepsilon,\delta')$-DP.
\end{defn}
The auxiliary lemma~\ref{lem:tight_d} is needed for Lemma~\ref{lem:maxrepr}.
For discrete valued distributions, it is given in~\cite[Lemma 1]{meiser2018tight} and another version of this result
using so called $f$-divergences is given in~\cite{barthe2013beyond}. We prove it here for for completeness, using our formalism.
In the proof, if $f_X$ and $f_Y$ are discrete valued distributions and if
$$
f_X(t) - \ee^\veps f_Y(t) = \sum\limits_i c_i \cdot \deltat{t_i}
$$
for some coefficients $c_i,t_i \in \mathbb{R}$, then $\max \{  f_X(t) - \ee^\veps f_Y(t) ,0  \} $ denotes
$$
\max \{  f_X(t) - \ee^\veps f_Y(t) ,0  \}  = \sum\limits_i \max \{c_i,0\} \cdot \deltat{t_i},
$$
and the set $S$ denotes
$$
S = \{ t \in \mathbb{R} \, : \, f_Y(t) \geq \ee^\veps f_X(t) \} = \mathbb{R} \setminus \{ t_i \, : \, c_i < 0 \}.
$$

\medskip

\begin{lem} \label{lem:tight_d}
$\mathcal{M}$ is tightly $(\veps,\delta)$-DP with
\begin{equation} \label{eq:max_eq}
\delta(\veps) = \max_{X \sim Y} \Bigg\{ \int\limits_\mathbb{R}  \max \{  f_X(t) - \ee^\veps f_Y(t) ,0  \} \, \dd t,  
\int\limits_\mathbb{R}  \max \{  f_Y(t) - \ee^\veps f_X(t) ,0  \} \, \dd t \Bigg\}.
\end{equation}
\begin{proof}
Assume $\mathcal{M}$ is tightly $(\veps,\delta)$-DP. Then, for every set $S \subset \mathbb{R}$ and for all $X \sim Y$:
\begin{equation*}
	\begin{aligned}
		\int\limits_S   f_X(t) - \ee^\veps f_Y(t) \, \dd t
		\leq  \int\limits_S  \max \{  f_X(t) - \ee^\veps f_Y(t) ,0  \} \, \dd t
		\leq  \int\limits_\mathbb{R}  \max \{  f_X(t) - \ee^\veps f_Y(t) ,0  \} \, \dd t.
	\end{aligned}
\end{equation*}	
We get an analogous bound for $\int_S f_Y(t) - \ee^\veps f_X(t) \, \dd t$.
Since $\mathcal{M}$ is tightly $(\veps,\delta)$-DP, by Definition~\ref{dfn:tightADP},
$$
\delta \leq \max \Bigg\{ \int\limits_\mathbb{R}  \max \{  f_X(t) - \ee^\veps f_Y(t) ,0  \} \, \dd t,  
\int\limits_\mathbb{R}  \max \{  f_Y(t) - \ee^\veps f_X(t) ,0  \} \, \dd t \Bigg\}.
$$
To show that the above inequality is tight, consider the set
$$
S = \{ t \in \mathbb{R} \, : \, f_X(t) \geq \ee^\veps f_Y(t) \}.
$$
Then, 
\begin{equation} \label{eq:caseAB}
	\begin{aligned}
		\int\limits_S  f_X(t) - \ee^\veps f_Y(t) \, \dd t &= \int\limits_S  \max \{ f_X(t) - \ee^\veps f_Y(t),0 \} \, \dd t \\
		&= \int\limits_\mathbb{R}  \max \{ f_X(t) - \ee^\veps f_Y(t),0 \} \, \dd t.
	\end{aligned}
\end{equation}
Next, consider the set $S = \{ t \in \mathbb{R} \, : \, f_Y(t) \geq \ee^\veps f_X(t) \}$.
Similarly,
\begin{equation} \label{eq:caseBA}
	\begin{aligned}
		\int\limits_S  f_Y(t) - \ee^\veps f_{X}(t) \, \dd t = \int\limits_\mathbb{R}  \max \{ f_Y(t) - \ee^\veps f_X(t),0 \} \, \dd t.
	\end{aligned}
\end{equation}
From \eqref{eq:caseAB} and \eqref{eq:caseBA} it follows that there exists a set $S \subset \mathbb{R}$ such that either
$$
\int_S f_X(t) \, \dd t = \ee^\varepsilon \int_S f_Y(t) \, \dd t  + \delta \quad \textrm{or} \quad
\int_S f_Y(t) \, \dd t = \ee^\varepsilon \int_S f_X(t) \, \dd t  + \delta
$$
for $\delta$ given by \eqref{eq:max_eq}. This shows that $\delta$ given by \eqref{eq:max_eq} is tight.
\end{proof}
\end{lem}

Recall from the main text that if $f_X$ and $f_Y$ are of the form (3.1), 
then the PLD distribution function is given by
\begin{equation} \label{eq:omega_pld2}
	\omega_{X/Y}(s) = \sum\limits_{{t_{X,i} = t_{Y,j} }}   a_{X,i} \cdot \deltas{s_i}, 
	\quad s_i = \log \left( \tfrac{a_{X,i}}{a_{Y,j}} \right).
\end{equation}
The following lemma gives an integral representation for the tight $\delta(\veps)$-bound
involving the distribution function of the PLD. 
For discrete valued distributions, it is originally given in~\cite[Lemma 5]{sommer2019privacy}.

\begin{lem}  \label{lem:maxrepr}
Let $\mathcal{M}$ be defined as above. 
$\mathcal{M}$  is tightly $(\varepsilon,\delta)$-DP for 
$$
\delta(\veps) = \max_{X \sim Y} \, \max \{ \delta_{X/Y}(\veps), \delta_{Y/X}(\veps) \}, 
$$
where
\begin{equation*}
	\begin{aligned}
		\delta_{X/Y}(\veps) &= \delta_{X/Y}(\infty) + \int\limits_\veps^\infty  (1-\ee^{\veps - s}) \, \omega_{X/Y}(s)   \, \dd s, \\
		\delta_{Y/X}(\veps) &= \delta_{Y/X}(\infty) + \int\limits_\veps^\infty  (1-\ee^{\veps - s}) \, \omega_{Y/X}(s)   \, \dd s, \\
		\delta_{X/Y}(\infty) 	
		&= 	\sum\limits_{ \{ t_i \, : \, \mathbb{P}( \mathcal{M}(X) = t_i) > 0, \, \mathbb{P}( \mathcal{M}(Y) = t_i) = 0 \} } 
			a_{X,i}, \\ 
		\delta_{Y/X}(\infty)  
		&= \sum\limits_{ \{ t_i \, : \, \mathbb{P}( \mathcal{M}(Y) = t_i) > 0, \, \mathbb{P}( \mathcal{M}(X) = t_i) = 0 \} } 
			a_{Y,i}. \\ 
	\end{aligned}
\end{equation*}
\begin{proof}
We directly find from the definition of $f_X$ and $f_Y$ and from the definition \eqref{eq:omega_pld2} that
\begin{equation*}
	\begin{aligned}
		\max \{ f_X(t) - \ee^\veps f_Y(t),0 \} &= \sum\limits_{ \{ t_i \, : \, \mathbb{P}( \mathcal{M}(X) = t_i) > 0, \, \mathbb{P}( \mathcal{M}(Y) = t_i) = 0 \} }  a_{X,i} \cdot 
		\deltat{ t_{X,i} } \\
		& + \sum\limits_{  t_{X,i} = t_{Y,j} }  \max \{ a_{X,i} - \ee^\veps a_{Y,j},0 \} \cdot \deltat{ t_{X,i} } \\
		& = \sum\limits_{ \{ t_i \, : \, \mathbb{P}( \mathcal{M}(X) = t_i) > 0, \, \mathbb{P}( \mathcal{M}(Y) = t_i) = 0 \} }  a_{X,i} \cdot \deltat{ t_{X,i} }  \\
	     & + \sum\limits_{  t_{X,i} = t_{Y,j} } a_{X,i} \max \{ ( 1  -  \ee^{\veps-s_i}),0 \} \cdot \deltat{ t_{X,i} },		
	\end{aligned}
\end{equation*}
and therefore
\begin{equation*}
	\begin{aligned}
\int\limits_\mathbb{R} \max \{ f_X(t) - \ee^\veps f_Y(t),0 \} \, \dd t
 & = \delta_{X/Y}(\infty) + \sum\limits_{ t_{X,i} = t_{Y,j} } a_{X,i} \max \{ ( 1  -  \ee^{\veps-s_i}),0 \} \\
 & = \delta_{X/Y}(\infty) + \int\limits_\veps^\infty  (1-\ee^{\veps - s}) \, \omega_{X/Y}(s)   \, \dd s.
	\end{aligned}
\end{equation*}
Analogously, we see that
\begin{equation*} 
\int\limits_\mathbb{R}  \max \{ f_Y(t) - \ee^\veps f_X(t), 0  \} \, \dd t
= \delta_{Y/X}(\infty) + \int\limits_\veps^\infty (1-\ee^{\veps - s}) \, \omega_{Y/X}(s)   \, \dd s.
\end{equation*}
The claim follows then from Lemma~\ref{lem:tight_d}. 
\end{proof}
\end{lem}

\subsection{Privacy Loss Distribution of Compositions}

The following theorem shows that the PLD distribution of discrete non-adaptive compositions is obtain using a discrete convolution.
We first recall the definition of convolution of two generalised functions as defined in the main text.
Suppose the distributions  $f_X$ and $f_Y$ are of the form 
\begin{equation*}
	\begin{aligned}
		f_X(t) &= \sum\nolimits_i a_{X,i} \cdot \deltat{t_{X,i}}, \\
		f_Y(t) &= \sum\nolimits_i a_{Y,i} \cdot \deltat{t_{Y,i}},
	\end{aligned}
\end{equation*}
where $t_{X,i},t_{Y,i} \in \mathbb{R}$ and $a_{X,i},a_{Y,i} \geq 0$.
We define the convolution $f_X * f_Y$ as
\begin{equation} \label{eq:conv_def2}
	(f_X * f_Y )(t) =  \sum_{i,j} a_{X,i} \, a_{Y,j} \cdot \deltat{t_{X,i} + t_{Y,j}}.  
\end{equation}
The result of the following theorem is originally given in~\cite[Thm.\;1]{sommer2019privacy}.
For completeness we give a proof using our notation with generalised probability density functions.

\begin{thm} \label{thm:convolutions} 
Let $f_X(t)$, $f_Y(t)$, $f_{X'}(t)$ and $f_{Y'}(t)$ denote the density functions of $\mathcal{M}(X)$, $\mathcal{M}(Y)$, 
$\mathcal{M'}(X)$ and $\mathcal{M'}(Y)$, respectively.
Denote by $\omega_{X/Y}$ the PLD distribution of $\mathcal{M}(X)$ over $\mathcal{M}(Y)$ and 
by $\omega_{X'/Y'}$ the PLD distribution of $\mathcal{M'}(X)$ over $\mathcal{M'}(Y)$.
Denote by $\widetilde{\omega}_{X/Y}$ the PLD of the non-adaptive composition $\mathcal{M} \circ \mathcal{M'} = (\mathcal{M},\mathcal{M'} )$. 
The density function of $\widetilde{\omega}_{X/Y}$ is given by
$$
\widetilde{\omega}_{X/Y}  = \omega_{X/Y} * \omega_{X'/Y'}.
$$
Moreover,
\begin{equation*}
	\begin{aligned}
		\widetilde{\delta}_{X/Y}(\infty) : & = \mathbb{P}( (\mathcal{M} \circ \mathcal{M'})(X) > 0, (\mathcal{M} \circ \mathcal{M'})(Y) = 0 ) \\
		&=  1 - \big(1 - \delta_{X/Y}(\infty) \big)\big(1 - \delta'_{X/Y}(\infty)\big),
	\end{aligned}
\end{equation*}
where 
$$
\delta_{X/Y}(\infty) = \mathbb{P}( \mathcal{M}(X) > 0, \mathcal{M}(Y) = 0 ), \quad 
\delta'_{X/Y}(\infty) = \mathbb{P}( \mathcal{M'}(X) > 0, \mathcal{M'}(Y) = 0 ).
$$
\begin{proof}
By definition of the privacy loss distribution, 
\begin{equation*}
	\begin{aligned}
\widetilde{\omega}_{X/Y}(s) & = \sum\limits_{(t_i,t_i') = (t_j,t_j')}  \mathbb{P}\big( (\mathcal{M} \circ \mathcal{M'})(X) = (t_i,t_i') \big)
\cdot \deltas{ \widetilde{s}_i }, \\ 
& \quad \quad \widetilde{s}_i = \log \left( \frac{ (\mathcal{M} \circ \mathcal{M'})(X) = (t_i,t_i') }
{ (\mathcal{M} \circ \mathcal{M'})(Y) = (t_j,t_j') } \right).
	\end{aligned}
\end{equation*}
Due to the independence of $\mathcal{M}$ and $\mathcal{M'}$,
\begin{equation} \label{eq:independence}
	\begin{aligned}
\mathbb{P}\big( \mathcal{M}(X) &=t_i, \, \mathcal{M'}(X) = t_i'\big) = \mathbb{P}\big( \mathcal{M}(X)=t_i\big) \, \mathbb{P}\big( \mathcal{M'}(X) = t_i'\big), \\
\mathbb{P}\big( \mathcal{M}(Y) &=t_j, \, \mathcal{M'}(Y) = t_j' \big) = \mathbb{P}\big( \mathcal{M}(Y)=t_j\big) \, \mathbb{P}\big( \mathcal{M'}(Y) = t_j'\big). \\
	\end{aligned}
\end{equation}
Therefore,
\begin{equation*}
	\begin{aligned}
		\log \left(  \frac{\mathbb{P}\big( \mathcal{M}(X)=t_i, \, \mathcal{M'}(X) = t_i'\big)}{\mathbb{P}\big( \mathcal{M}(Y)=t_j, \, \mathcal{M'}(Y) = t_j'\big) }  \right) 
		= \log \left( \frac{\mathbb{P}\big( \mathcal{M}(X)=t_i\big)}{ \mathbb{P}\big( \mathcal{M}(Y)=t_j\big) }    \right) + 
		\log \left( \frac{ \mathbb{P}\big( \mathcal{M'}(X) = t_i'\big) }{ \mathbb{P}\big( \mathcal{M'}(Y) = t_j'\big) }    \right).
	\end{aligned}
\end{equation*}
and
\begin{equation} \label{eq:conv_last_step}
	\begin{aligned}
\widetilde{\omega}_{X/Y}(s) = \sum\limits_{(t_i,t_i') = (t_j,t_j')}  \mathbb{P}\big( \mathcal{M}(X)=t_i\big) \, \mathbb{P}\big( \mathcal{M'}(X) = t_i'\big)
\cdot \deltas{s_i + s_i'}, 
	\end{aligned}
\end{equation}
where
$$ 
s_i = \log \left( \frac{ \mathbb{P}\big( \mathcal{M}(X) = t_i \big) }{ \mathbb{P}\big( \mathcal{M}(Y) = t_j \big) } \right),
\quad s_i' = \log \left( \frac{ \mathbb{P}\big( \mathcal{M'}(X) = t_i' \big) }{ \mathbb{P}\big( \mathcal{M'}(Y) = t_j' \big) } \right).
$$
We see from \eqref{eq:conv_last_step} that $\widetilde{\omega}_{X/Y}  = \omega_{X/Y} * \omega_{X'/Y'}$ with convolution defined in \eqref{eq:conv_def2}.
The expression for $\widetilde{\delta}_{X/Y}(\infty)$ follows directly from its definition and from the independence of the mechanisms \eqref{eq:independence}.
\end{proof}
\end{thm}
Theorem~\ref{thm:convolutions} directly gives the following representation for tight $\delta(\veps)$ of compositions.
\begin{cor}
Consider $k$ consecutive applications of a mechanism $\mathcal{M}$. Let $\veps > 0$. The composition is tightly $(\veps,\delta)$-DP for $\delta$ given by
$$
\delta(\veps) = \max_{X \sim Y} \max \{ \delta_{X/Y}(\veps), \delta_{Y/X}(\veps) \},
$$ 
where
\begin{equation*} 
	\begin{aligned}
		\delta_{X/Y}(\veps) =
	1 - (1-\delta_{X/Y}(\infty))^k +	\int\limits_\veps^\infty (1 - \ee^{\veps - s})\left( \omega_{X/Y} *^k  \omega_{X/Y}\right) (s)  \, \dd s,
	\end{aligned}
\end{equation*}
where $(\omega_{X/Y} *^k \omega_{X/Y}) (s)$ denotes the density function 
obtained by convolving $\omega_{X/Y}$ by itself $k$ times (an analogous formula holds for $\delta_{Y/X}(\veps)$).
\end{cor}

\section{Proofs for the Results of Section 4}

\subsection{Grid Approximation}

Recall from Section 4 of the main text: we place the PLD distribution on a grid $X_n = \{x_0,\ldots,x_{n-1}\}$, $n \in \mathbb{Z}^+$, where
\begin{equation} \label{eq:grid}
	x_i = -L + i \Delta x, \quad  \Delta x = 2L/n. 
\end{equation}
Suppose the distribution $\omega$ of the PLD is of the form 
\begin{equation} \label{eq:omega_plain}
	\omega(s) = \sum_{i=0}^{n-1} a_i \cdot \deltas{s_i},
\end{equation}	
where $a_i \geq 0$ and $-L \leq s_i \leq L - \Delta x$, $0 \leq i \leq n-1$. 
We define the grid approximations
\begin{equation} \label{eq:omegaRL}
	\begin{aligned}
		\omega^\mathrm{L}(s) &= \sum\limits_{i=0}^{n-1} a_i \cdot \deltas{s_i^\mathrm{L}},     \quad s_i^\mathrm{L} = \sup \{  x \in X_n \, : \, s_i \geq x \}, \\
		\omega^\mathrm{R}(s) &= \sum\limits_{i=0}^{n-1} a_i \cdot \deltas{s_i^\mathrm{R}},     \quad s_i^\mathrm{R} = \inf \{  x \in X_n \, : \, s_i \leq x \}.
	\end{aligned}
\end{equation}

\begin{lem} \label{lem:deltaineq}
Let $\delta(\veps)$ be given by the integral formula of Lemma~\ref{lem:maxrepr} 
and let $\delta^\mathrm{L}(\veps)$ and $\delta^\mathrm{R}(\veps)$ be defined analogously by $\omega^\mathrm{L}$ and $\omega^\mathrm{R}$.
Then for all $\veps>0$ we have
\begin{equation} \label{eq:delta_ineq}
	\delta^\mathrm{L}(\veps) \leq \delta(\veps) \leq \delta^\mathrm{R}(\veps).
\end{equation}
\begin{proof}
The claim follows from the definition \eqref{eq:omegaRL} and from
the fact that $(1-\ee^{\veps - s})$ is a monotonously increasing function of $s$.
\end{proof}
\end{lem}

\begin{cor}
Lemma~\ref{lem:deltaineq} directly generalises to convolutions. Namely, if 
$$
(\omega *^k \omega)(s) = \sum_i a_i \cdot \deltas{s_i}
$$
for some coefficients $a_i \geq 0$, $s_i \in \mathbb{R}$,
then from the definition \eqref{eq:conv_def2} it follows that
$$
(\omega^\mathrm{L} *^k \omega^\mathrm{L})(s) = \sum_i a_i \cdot \deltas{s_i^\mathrm{L}}
$$
for some $s_i^\mathrm{L}$ such that $s_i^\mathrm{L} \leq s_i$ for all $i$. And similarly, then
$$
(\omega^\mathrm{R} *^k \omega^\mathrm{R})(s) = \sum_i a_i \cdot \deltas{s_i^\mathrm{R}}
$$
for some $s_i^\mathrm{R}$ such that $s_i^\mathrm{R} \geq s_i$ for all $i$.
And since $(1-\ee^{\veps - s})$ is a monotonously increasing function of $s$ for $s \geq \veps$,
the inequality \eqref{eq:delta_ineq} holds also in case $\delta(\veps)$, $\delta^\mathrm{L}(\veps)$ and $\delta^\mathrm{R}(\veps)$ is determined by 
$\omega *^k \omega$,  $\omega^\mathrm{L} *^k \omega^\mathrm{L}$ and $\omega^\mathrm{R} *^k \omega^\mathrm{R}$, respectively.
\end{cor}

The following bounds for the moment generating functions will be used in the error analysis.

\begin{lem} \label{lem:mgfineq}
Let $\omega, \, \omega^\mathrm{R}$ and $\omega^\mathrm{L}$ be defined as in \eqref{eq:omega_plain} and \eqref{eq:omegaRL} and let $0 < \lambda < (\Delta x)^{-1}$. Then
\begin{equation} \label{eq:bE1}
	\mathbb{E} [ \ee^{ \lambda \omega^\mathrm{L}} ] \leq \mathbb{E} [ \ee^{ \lambda \omega} ], \quad \quad
	\mathbb{E} [ \ee^{  - \lambda \omega^\mathrm{L}} ] \leq \tfrac{1}{1-\lambda \Delta x} \mathbb{E} [ \ee^{ - \lambda \omega} ]
\end{equation}	
and
\begin{equation} \label{eq:bE2}
	\mathbb{E} [ \ee^{ \lambda \omega^\mathrm{R}} ] \leq \tfrac{1}{1-\lambda \Delta x} \mathbb{E} [ \ee^{ \lambda \omega } ], \quad \quad
	\mathbb{E} [ \ee^{ - \lambda \omega^\mathrm{R}} ] \leq \mathbb{E} [ \ee^{ - \lambda \omega } ].
\end{equation}
\begin{proof}
The condition $\mathbb{E} [ \ee^{ \lambda \omega^\mathrm{L}} ] \leq \mathbb{E} [ \ee^{ \lambda \omega} ]$ 
follows directly from the definition \eqref{eq:omegaRL}:
$$
\mathbb{E} [ \ee^{ \lambda \omega^\mathrm{L}} ] = 
\sum\limits_{i=0}^{n-1} a_i   \ee^{\lambda s_i^\mathrm{L} } \leq \sum\limits_{i=0}^{n-1} a_i   \ee^{\lambda s_i }
= \mathbb{E} [ \ee^{ \lambda \omega } ],
$$ 
since $s_i^\mathrm{L} \leq s_i$ for all $0 \leq i \leq n-1$. The proof for the condition
$\mathbb{E} [ \ee^{ - \lambda \omega^\mathrm{R}} ] \leq \mathbb{E} [ \ee^{ - \lambda \omega } ]$ goes similarly.

Using the Lipschitz continuity of the exponential function, we see that
\begin{equation*}
	\begin{aligned}
		\mathbb{E} [ \ee^{ \lambda \omega^\mathrm{R}} ] - \mathbb{E} [ \ee^{ \lambda \omega} ] &=
		\sum\limits_{i=0}^{n-1} a_i \big(  \ee^{\lambda s_i^\mathrm{R} } -  \ee^{\lambda s_i }  \big)  \\
		 &\leq   \sum\limits_{i=0}^{n-1} a_i \lambda  \abs{s_i^\mathrm{R} - s_i } \ee^{\lambda s_i^\mathrm{R} } \\
		 &\leq \lambda \Delta x \sum\limits_{i=0}^{n-1} a_i \ee^{\lambda s_i^\mathrm{R} } 
		 = \lambda \Delta x \, \mathbb{E} [ \ee^{ \lambda \omega^\mathrm{R}} ].
	\end{aligned}
\end{equation*}
Thus
$$
(1-\lambda \Delta x) \, \mathbb{E} [ \ee^{ \lambda \omega^\mathrm{R}} ] \leq \mathbb{E} [ \ee^{ \lambda \omega} ]
$$
from which the condition
$\mathbb{E} [ \ee^{ \lambda \omega^\mathrm{R}} ] \leq \tfrac{1}{1-\lambda \Delta x} \, \mathbb{E} [ \ee^{ \lambda \omega } ]$
follows. The proof for the condition
$\mathbb{E} [ \ee^{  - \lambda \omega^\mathrm{L}} ] \leq \tfrac{1}{1-\lambda \Delta x} \, \mathbb{E} [ \ee^{ - \lambda \omega} ]$
goes similarly.
\end{proof}
\end{lem}

\subsection{FFT Evaluation for Truncated Convolutions of Periodic Distributions}

We next prove the lemma showing that the truncated convolutions of periodic distributions
can be evaluated using FFT. 
Suppose $\omega$ is defined on $X_n$ such that 
\begin{equation} \label{eq:omega}
	\omega(s) = \sum\limits_{i=0}^{n-1} a_i \cdot \deltas{s_i},
\end{equation}
where $a_i \geq 0$ and $s_i = i \Delta x$. 
The convolutions can then be written as 
\begin{equation*}
	\begin{aligned}
	(\omega * \omega )(s)  =  \sum\limits_{  i,j} a_i a_j \cdot \deltas{s_i+s_j} 
	 = \sum\limits_i \Big(\sum\limits_j a_j a_{i-j} \Big) \cdot \deltas{s_i}.
\end{aligned}
\end{equation*}
We define $\widetilde{\omega}$ to be a $2 L$-periodic extension of $\omega$ such that  
$$
\widetilde{\omega}(s) = \sum\limits_{m \in \mathbb{Z}} \, \sum\limits_i a_i \cdot \deltas{s_i+m \cdot 2 L}.
$$
In case the distribution $\omega$ is defined on an equidistant grid, FFT can be used to evaluate the
approximation $\widetilde{\omega} \circledast \widetilde{\omega} $:
\begin{lem} \label{lem:fft}
Let $\omega$ be of the form \eqref{eq:omega}, such that
$n$ is even and $s_i = -L + i \Delta x$, $0 \leq i \leq n-1$. 
Define
$$
\boldsymbol{a} = \begin{bmatrix} a_0 & \ldots & a_{n-1} \end{bmatrix}^\mathrm{T} \quad \textrm{and} \quad
 D = \begin{bsmallmatrix} 0 & I_{n/2} \\ I_{n/2} & 0 \end{bsmallmatrix} \in \mathbb{R}^{n \times n}.
$$
Then, 
$$
(\widetilde{\omega} \circledast^k \widetilde{\omega} )(s) = \sum\limits_{i=0}^{n-1} b_i^k \cdot \deltas{s_i},
$$
where
$$
b_i^k = \left[D \, \mathcal{F}^{-1} \big(\mathcal{F}( D \boldsymbol{a} )^{ \odot k}   \big) \right]_i,
$$
and $^{ \odot k}$ denotes the elementwise power of vectors.
\begin{proof}
Assume $n$ is even and $s_i = -L + i \Delta x$, $0 \leq i \leq n-1$. 
From the the truncation and periodisation it follows that
$\widetilde{\omega} \circledast \widetilde{\omega}$ is of the form
\begin{equation} \label{eq:b_i}
	(\widetilde{\omega} \circledast \widetilde{\omega} )(s) = \sum\limits_{i=0}^{n-1} b_i \cdot \deltas{s_i},
	\quad \quad b_i = \sum\limits_{j = n/2}^{3n/2-1} a_j \, a_{i - j } \, \textrm{ (indices modulo $n$)}.
\end{equation}
Denoting $\boldsymbol{\widetilde{a}} = D \boldsymbol{a}$,
we see that the coefficients $b_i$ in \eqref{eq:b_i} are given by the expression
$$
b_{i+n/2} = \sum\limits_{j = 0}^{n-1} \widetilde{a}_j \, \widetilde{a}_{i - j } \, \textrm{ (indices modulo $n$)},
$$
to which we can apply DFT and the convolution theorem \cite{stockham1966}. I.e., when $0 \leq i \leq n-1$,
\begin{equation} \label{eq:FinvF}
	b_{i+n/2} = \left[ \mathcal{F}^{-1} \big(\mathcal{F}( \boldsymbol{\widetilde{a}} ) \odot  \mathcal{F} ( \boldsymbol{\widetilde{a}} )  \big) \right]_i 
	= \left[ \mathcal{F}^{-1} \big(\mathcal{F}( D \boldsymbol{a} ) \odot  \mathcal{F} ( D \boldsymbol{a} )  \big) \right]_i,  \, \textrm{ (indices modulo $n$)}
\end{equation}
where $\odot$ denotes the elementwise product of vectors. From \eqref{eq:FinvF} we find that
$$
	b_i = \left[ D \mathcal{F}^{-1} \big(\mathcal{F}( D \boldsymbol{a} ) \odot  \mathcal{F} ( D \boldsymbol{a} )  \big) \right]_i,  \, \textrm{ (indices modulo $n$)}.
$$
By induction this generalises to $k$-fold compositions and we arrive at the claim.
\end{proof}
\end{lem}

\section{Proof of Theorem 10} \label{sec:err_est_a} 


We next prove step by step the main theorem, i.e., Theorem 10 of the main text. We start by splitting the error
induced by Algorithm 1 into three terms.

\begin{lem} \label{thm:total_error}
Let $\omega$ be a generalised distribution and denote by $\widetilde{\delta}(\veps)$ the result of Algorithm 1.
Total error of the approximation can be split as follows:
\begin{equation*} 
	\begin{aligned}
   \abs{\int\limits_\veps^\infty (1 - \ee^{\veps - s})(\omega *^k \omega ) (s)  \, \dd s -  \widetilde{\delta}(\veps)} 
\leq  I_1(L) + I_2(L) + I_3(L),
	\end{aligned}
\end{equation*}
where
\begin{equation*} 
\begin{aligned}
 I_1(L)  & =  \int\limits_L^\infty (\omega *^k \omega ) (s)  \, \dd s, \\
 I_2(L)  &=  \int\limits_{\veps}^L (\omega *^k \omega - \omega \circledast^k \omega)(s) \, \dd s, \\ 
 I_3(L)  &=  \int\limits_\veps^L\abs{(\omega \circledast^k \omega - \widetilde{\omega} \circledast^k \widetilde{\omega})(s)} \, \dd s,  
	\end{aligned}
\end{equation*}
where, for a generalised density function of the form $\sum\nolimits_i a_i \cdot \deltas{s_i} $, the absolute value denotes
$$
\abs{ \sum\limits_i a_i \cdot \deltas{s_i} } =  \sum\limits_i \abs{a_i} \cdot \deltas{s_i}.
$$
\begin{proof}
By adding and subtracting terms and using the triangle inequality, we get
\begin{equation} \label{eq:total_error}
	\begin{aligned}
   \int\limits_\veps^\infty (1 - \ee^{\veps - s})(\omega *^k \omega ) (s)  \, \dd s -  \widetilde{\delta}(\veps) 
& =  	  \int\limits_\veps^\infty (1 - \ee^{\veps - s})(\omega *^k \omega ) (s)  \, \dd s -
		\int\limits_\veps^L (1 - \ee^{\veps - s})(\omega *^k \omega ) (s)  \, \dd s \\ 
&	\quad + \int\limits_\veps^L (1 - \ee^{\veps - s})(\omega *^k \omega ) (s)  \, \dd s -
		\int\limits_\veps^L (1 - \ee^{\veps - s})(\widetilde{\omega} \circledast^k \widetilde{\omega} ) (s)  \, \dd s. 
	\end{aligned}
\end{equation}
Since $0 \leq (1 - \ee^{\veps - s}) < 1$ for all $s \geq \veps$,
we have for the first term on the right hand side of \eqref{eq:total_error}:
\begin{equation} \label{eq:tail}
0 \leq \int\limits_\veps^\infty (1 - \ee^{\veps - s})(\omega *^k \omega ) (s)  \, \dd s -
		\int\limits_\veps^L (1 - \ee^{\veps - s})(\omega *^k \omega ) (s)  \, \dd s  \leq 
		\int\limits_L^\infty (\omega *^k \omega ) (s)  \, \dd s.
\end{equation}
Similarly, adding and subtracting $\int\limits_\veps^L (1 - \ee^{\veps - s})(\omega \circledast^k \omega ) (s)  \, \dd s$ the second term 
on the right hand side of \eqref{eq:total_error}, we find that
$$
\abs{\int\limits_\veps^L (1 - \ee^{\veps - s})(\omega *^k \omega ) (s)  \, \dd s -
		\int\limits_\veps^L (1 - \ee^{\veps - s})(\widetilde{\omega} \circledast^k \widetilde{\omega} ) (s)  \, \dd s }
		\leq I_2(L) + I_3(L)
$$
which shows the claim.
\end{proof}
\end{lem}
We next consider separately each of the three terms stated in Theorem~\ref{thm:total_error}.
Each of them are bounded using the Chernoff bound~\cite{wainwright2019}
\begin{equation*} 
\mathbb{P}[ X \geq t] = \mathbb{P}[ \ee^{\lambda X} \geq \ee^{\lambda t} ] \leq \frac{ \mathbb{E}[ \ee^{\lambda X} ] }{\ee^{\lambda t}}
\end{equation*}
which holds for any random variable $X$ and for all $\lambda > 0$.
If $\omega$ is of the form 
$$
\omega(s) = \sum_{i=0}^{n-1} a_i \cdot \deltas{s_i}, \quad s_i = \log \left( \frac{a_{X,i}}{a_{Y,i}}   \right),
$$ 
where $a_{X,i},a_{Y,i} \geq 0$, $s_i \in \mathbb{R}$, $0 \leq i \leq n-1$, the moment generating function is given by
\begin{equation} \label{eq:pld_lmf}
	\begin{aligned}
		\mathbb{E} [\ee^{\lambda \omega_{X/Y} }] &= \int\limits_{-\infty}^\infty \ee^{\lambda s} \omega(s) \, \dd s 
		= \sum\limits_{i=1}^n  \ee^{\lambda s_i} \cdot a_{X,i} 
		= \sum\limits_{i=1}^n  \left( \frac{a_{X,i}}{a_{Y,i}}  \right)^\lambda a_{X,i}.
	\end{aligned}
\end{equation}

\subsection{Tail Bound for the Convolved PLDs}

Denote $S_k := \sum_{i=1}^k \omega_i$, where $\omega_i$ denotes the PLD random variable of the $i$th mechanism.
Since $\omega_i$'s are independent, $ \mathbb{E} [ \ee^{\lambda S_k}  ]  = \prod_{i=1}^k \mathbb{E} [ \ee^{\lambda \omega_i}  ] $
and the Chernoff bound shows that for any $\lambda > 0$
$$
\int_L^\infty  ( \omega \ast^k \omega)(s) \, \dd s   = \mathbb{P}[ S_k \geq L ] \leq \prod_{i=1}^k \mathbb{E} [ \ee^{\lambda \omega_i}  ] \,  \ee^{- \lambda L}.
$$
If $\omega_i$'s are i.i.d. and distributed as $\omega$, and if $\alpha(\lambda) = \log( \mathbb{E} [\ee^{\lambda \omega }] )$, then
\begin{equation} \label{eq:chernoff}
I_1(L) = 	\int\limits_L^\infty  ( \omega \ast^k \omega)(s) \, \dd s  \leq \ee^{k \alpha(\lambda)}  \ee^{- \lambda L}. 
\end{equation}

\subsection{Error Arising from the Periodisation} \label{subsec:third_approx}

We define $\alpha^+(\lambda)$ and $\alpha^-(\lambda)$ via the moment generating function 
of the PLD as
\begin{equation} \label{eq:conn}
	\begin{aligned}
	\alpha^+(\lambda) = \log (	\mathbb{E} [\ee^{\lambda \omega }] ) \quad \textrm{and}
	\quad \alpha^-(\lambda) = \log (	\mathbb{E} [\ee^{ - \lambda \omega }] ).
	\end{aligned}
\end{equation}
Using the Chernoff bound, the required error bounds can be obtained using $\alpha^+(\lambda)$ and $\alpha^-(\lambda)$.

\begin{lem} \label{lem:period}
Let $\omega$ be defined as above and suppose $s_i \in [-L,L]$ for all $0\leq i \leq n-1$. Then, 
\begin{equation*}
	\begin{aligned}
 I_3(L) = \int\limits_\veps^L\abs{(\omega \circledast^k \omega - \widetilde{\omega} \circledast^k \widetilde{\omega})(s)} \, \dd s  
 \leq \big(  \ee^{k \alpha^+(\lambda)} + \ee^{k \alpha^-(\lambda)}  \big) \frac{ \ee^{- L \lambda} }{ 1 - \ee^{- L \lambda}}.
	\end{aligned}
\end{equation*}	
\begin{proof}

Let $\omega$ and its $2L$-periodic continuation $\widetilde{\omega}(s)$ be of the form
$$
\omega(s) = \sum_i a_i \cdot \deltas{s_i} \quad \textrm{and} \quad 
\widetilde{\omega}(s) = \sum_i \widetilde{a}_i \cdot \deltas{s_i}
$$
for some $a_i,\widetilde{a}_i \geq 0$, $s_i = i \Delta x$. 	By definition of the truncated convolution $\circledast$ (see the main text),
\begin{equation*}
	\begin{aligned}
	(\widetilde{\omega} \circledast^k \widetilde{\omega})(s) & = 
	\sum\limits_{-L \leq s_{j_1} < L} \widetilde{a}_{j_1} \sum\limits_{-L \leq s_{j_2} < L} \widetilde{a}_{j_2} 
	\ldots \sum\limits_{-L \leq s_{j_{k-1}} < L} \widetilde{a}_{j_{k-1}} 
	\sum\limits_i  \widetilde{a}_{i - j_1 - \ldots - j_{k-1}} \cdot \deltas{s_i}  \\
	& = \sum\limits_{-L \leq s_{j_1} < L} a_{j_1} \sum\limits_{-L \leq s_{j_2} < L} a_{j_2} 
		\ldots \sum\limits_{-L \leq s_{j_{k-1}} < L} a_{j_{k-1}} 
		\sum\limits_i  \widetilde{a}_{i - j_1 - \ldots - j_{k-1}} \cdot \deltas{s_i} \\
	& = \sum\limits_{j_1} a_{j_1} \sum\limits_{j_2}  a_{j_2} 
		\ldots \sum\limits_{j_{k-1}}  a_{j_{k-1}} 
		\sum\limits_i  \widetilde{a}_{i - j_1 - \ldots - j_{k-1}} \cdot \deltas{s_i},
	\end{aligned}
\end{equation*}	
since $\widetilde{a}_i = a_i$ for all $i$ such that $-L \leq s_i < L$.
Furthermore,
\begin{equation*}
	\begin{aligned}
	(\omega \circledast^k \omega)(s)
	& = \sum\limits_{-L \leq s_{j_1} < L} a_{j_1} \sum\limits_{-L \leq s_{j_2} < L} a_{j_2} 
		\ldots \sum\limits_{-L \leq s_{j_{k-1}} < L} a_{j_{k-1}} 
		\sum\limits_i  a_{i - j_1 - \ldots - j_{k-1}} \cdot \deltas{s_i} \\
	& = \sum\limits_{j_1} a_{j_1} \sum\limits_{j_2}  a_{j_2} 
		\ldots \sum\limits_{j_{k-1}}  a_{j_{k-1}} 
		\sum\limits_i  a_{i - j_1 - \ldots - j_{k-1}} \cdot \deltas{s_i}.
	\end{aligned}
\end{equation*}	
Thus
\begin{equation} \label{eq:C31}
	\begin{aligned}
	(\widetilde{\omega} \circledast^k \widetilde{\omega} - \omega \circledast^k \omega)(s)	= \sum\limits_{j_1}  a_{j_1} \sum\limits_{j_2}  a_{j_2} 
		\ldots \sum\limits_{j_{k-1}} a_{j_{k-1}} 
		\sum\limits_i  \widehat{a}_{i - j_1 - \ldots - j_{k-1}} \cdot \deltas{s_i},
	\end{aligned}
\end{equation}	
where 
\begin{equation} \label{eq:C32}
	\widehat{a}_{i} = \widetilde{a}_i - a_i = \begin{cases}
		 0 , &\text{ if } -L \leq s_i < L, \\
				a_{i \, \textrm{mod} \, n}, &\text{ else. }
	\end{cases}
\end{equation}
From \eqref{eq:C31} we see that
\begin{equation} \label{eq:C34}
	\begin{aligned}
 \int\limits_\veps^L\abs{(\omega \circledast^k \omega - \widetilde{\omega} \circledast^k \widetilde{\omega})(s)} \, \dd s 
 & \leq \int\limits_{\mathbb{R}} \abs{(\omega \circledast^k \omega - \widetilde{\omega} \circledast^k \widetilde{\omega})(s)} \, \dd s \\
& = \int\limits_{\mathbb{R}} \sum\limits_{j_1} a_{j_1} \sum\limits_{j_2} a_{j_2} 
		\ldots \sum\limits_{j_{k-1}} a_{j_{k-1}} 
		\sum\limits_i  \widehat{a}_{i - j_1 - \ldots - j_{k-1}} \cdot \deltas{s_i} \, \dd s \\
&  = \sum\limits_{j_1} a_{j_1} \sum\limits_{j_2} a_{j_2} 
		\ldots \sum\limits_{j_{k-1}} a_{j_{k-1}} 
		\sum\limits_i  \widehat{a}_{i - j_1 - \ldots - j_{k-1}}.
	\end{aligned}
\end{equation}
From \eqref{eq:C32} we see that
\begin{equation} \label{eq:C35}
	\begin{aligned}
 \sum\limits_{j_1} a_{j_1} \sum\limits_{j_2} a_{j_2} 
		\ldots \sum\limits_{j_{k-1}} a_{j_{k-1}} 
		\sum\limits_i  \widehat{a}_{i - j_1 - \ldots - j_{k-1}} 
		= & \sum\limits_{n \in \mathbb{Z}\setminus \{0\}} \mathbb{P} \big( (2 n - 1) L \leq  \omega *^k \omega < (2n+1) L \big) \\
		= & \sum\limits_{n \in \mathbb{Z}^-} \mathbb{P} \big( (2 n - 1) L \leq  \omega *^k \omega < (2n+1) L \big) \\
		& \quad \quad + \sum\limits_{n \in \mathbb{Z}^+} \mathbb{P} \big( (2 n - 1) L \leq  \omega *^k \omega < (2n+1) L \big) \\
		\leq & \sum\limits_{n \in \mathbb{Z}^-} \mathbb{P} \big( \omega *^k \omega \leq (2n+1) L \big)
		+ \sum\limits_{n \in \mathbb{Z}^+} \mathbb{P} \big(  \omega *^k \omega  \geq (2 n - 1) L \big). \\
	\end{aligned}
\end{equation}
We also see that
$$
\sum\limits_{n \in \mathbb{Z}^-} \mathbb{P} \big( \omega *^k \omega \leq (2n+1) L \big) 
= \sum\limits_{n \in \mathbb{Z}^+} \mathbb{P} \big( (-\omega) *^k (-\omega) \geq (2n-1) L \big).
$$
Using the bounds \eqref{eq:C34}, \eqref{eq:C35} and the Chernoff bound \eqref{eq:chernoff}, we find that for all
$\lambda > 0$
\begin{equation*}
	\begin{aligned}
		\int\limits_\veps^L\abs{(\omega *^k \omega - \widetilde{\omega} \circledast^k \widetilde{\omega})(s)} \, \dd s 
		 & \leq  \sum\limits_{\ell=1}^\infty \ee^{k \alpha^+(\lambda)} \ee^{- \ell L \lambda} + \ee^{k \alpha^-(\lambda)} \ee^{- \ell L \lambda}  \\
		 & = \big(  \ee^{k \alpha^+(\lambda)} + \ee^{k \alpha^-(\lambda)}  \big) \frac{ \ee^{- L \lambda} }{ 1 - \ee^{- L \lambda}}.
	\end{aligned}
\end{equation*}
\end{proof}
\end{lem}

\subsection{Error Arising from the Truncation of the Convolution Integrals } \label{subsec:trunc_analysis}

Next, assume that the generalised distribution $\omega$ of the PLD is of the form 
$$
\omega(s) = \sum\limits_i a_i \cdot \deltas{s_i},
$$
where $a_i \geq 0$ and $s_i = i \Delta x$. 

The following lemma gives a bound for the truncation error
$\int\limits_\veps^L \abs{(\omega *^k \omega - \omega \circledast^k \omega)(s)} \, \dd s$ in terms of the moment generating function of $\omega$.
Notice that this result applies also for the case where the support of the PLD distribution are outside of the interval $[-L,L]$.
\begin{lem} \label{lem:trunc}
Let $\omega$ be defined as above. For all $\lambda > 0$,
$$
I_2(L)= \int\limits_{\veps}^L (\omega *^k \omega - \omega \circledast^k \omega)(s) \, \dd s \leq
		\bigg( \frac{\ee^{k \alpha^+(\lambda)}-\ee^{ \alpha^+(\lambda)}}{\ee^{\alpha^+(\lambda)}-1}
		+ \frac{\ee^{ \alpha^-(\lambda)}-\ee^{k \alpha^-(\lambda)}}{1-\ee^{\alpha^-(\lambda)}} \bigg) \, \ee^{-L \lambda}.
$$
\begin{proof}
By adding and subtracting $(\omega *^k \omega) \circledast \omega $ , we may write
\begin{equation} \label{eq:C21}
	\omega *^k \omega - \omega \circledast^k \omega =
	(\omega *^{k-1} \omega) * \omega - (\omega *^{k-1} \omega) \circledast \omega
	+  ( \omega *^{k-1} \omega - \omega \circledast^{k-1} \omega ) \circledast \omega.
\end{equation}
Let $\ell \in \mathbb{Z}^+$. Let $\omega$ be of the form $\omega(s) = \sum_i a_i \cdot \deltas{s_i}$
and let the convolution $\omega *^\ell \omega$ be of the form $(\omega *^\ell \omega)(s) = \sum_i c_i \cdot \deltas{s_i}$
for some $a_i,c_i \geq 0$, $s_i = i \Delta x$. 
From the definition of the operators $*$ and $\circledast$ it follows that
\begin{equation*}
	\begin{aligned}
		 \big( (\omega *^\ell \omega) * \omega - (\omega *^\ell \omega) \circledast \omega\big) (s)
		= & \sum\limits_i \Big( \sum\limits_j  c_j a_{i-j}   \Big) \cdot \deltas{s_i}
		- \sum\limits_i \Big( \sum\limits_{-L \leq s_j < L}  c_j a_{i-j}   \Big) \cdot \deltas{s_i} \\
		= & \sum\limits_i \Big( \sum\limits_{ s_j < -L, \, s_j \geq L}  c_j a_{i-j}   \Big) \cdot \deltas{s_i}.
	\end{aligned}
\end{equation*}
Therefore
\begin{equation} \label{eq:C22}
	\begin{aligned}
 \int\limits_{\mathbb{R}} \big(  (\omega *^\ell \omega) * \omega - (\omega *^\ell \omega) \circledast \omega\big) (s) \, \dd s
	= & \int\limits_{\mathbb{R}} \sum\limits_i \Big( \sum\limits_{ s_j < -L, \, s_j \geq L}  c_j a_{i-j}   \Big) \cdot \deltas{s_i} \, \dd s \\
	= & \sum\limits_{ s_j < -L, \, s_j \geq L}  c_j  \int\limits_{\mathbb{R}}  \sum\limits_i  a_{i-j}  \cdot \deltas{s_i} \, \dd s \\
	= & \sum\limits_{ s_j < -L, \, s_j \geq L}  c_j \\
    = & \, \mathbb{P} \Big( \omega *^\ell \omega < -L \Big) +  \mathbb{P} \Big( \omega *^\ell \omega \geq L \Big)  \\
	\leq & \, \ee^{\ell \alpha^+(\lambda)} \ee^{-L \lambda} + \ee^{\ell \alpha^-(\lambda)} \ee^{-L \lambda}
	\end{aligned}
\end{equation}
for all $\lambda>0$. The last inequality follows from the Chernoff bound.
Similarly, let $\omega *^\ell \omega - \omega \circledast^\ell \omega$
be of the form 
$$
(\omega *^\ell \omega - \omega \circledast^\ell \omega)(s) = \sum_i \widetilde{c}_i \cdot \deltas{s_i}
$$
for some $\widetilde{c}_i \geq 0$, $s_i = i \Delta x$. Then
\begin{equation} \label{eq:C23}
	\begin{aligned}
	 \int\limits_{\mathbb{R}} \big( ( \omega *^\ell \omega - \omega \circledast^\ell \omega ) \circledast \omega \big) (s) \, \dd s
	= & \int\limits_{\mathbb{R}} \sum\limits_i \Big( \sum\limits_{-L \leq s_j < L}  \widetilde{c}_j a_{i-j}   \Big) \cdot \deltas{s_i} \, \dd s \\
	= & \sum\limits_{-L \leq s_j < L}  \widetilde{c}_j \int\limits_{\mathbb{R}} \sum\limits_i a_{i-j}   \cdot \deltas{s_i} \, \dd s \\
	\leq & \sum\limits_{-L \leq s_j < L}  \widetilde{c}_j \\
	\leq & \int\limits_{\mathbb{R}} ( \omega *^\ell \omega - \omega \circledast^\ell \omega )(s) \, \dd s.
	\end{aligned}
\end{equation}
Using \eqref{eq:C21}, \eqref{eq:C22} and \eqref{eq:C23}, we see that for all $\lambda > 0$,
\begin{equation} \label{eq:C24}
	\begin{aligned}
		\int\limits_{\veps}^L (\omega *^k \omega - \omega \circledast^k \omega)(s) \, \dd s
		\leq & \, \, \int\limits_{\mathbb{R}} (\omega *^k \omega - \omega \circledast^k \omega)(s) \, \dd s  \\
		\leq & \, \,  \ee^{ (k-1) \alpha^+(\lambda)} \ee^{-L \lambda} + \ee^{ (k-1) \alpha^-(\lambda)}  \ee^{-L \lambda} 
		  +		\int\limits_{\mathbb{R}} ( \omega *^{k-1} \omega - \omega \circledast^{k-1} \omega )(s) \, \dd s.
	\end{aligned}
\end{equation}
Using \eqref{eq:C24} recursively, we see that for all $\lambda > 0$,
\begin{equation*}
	\begin{aligned}
		\int\limits_{\veps}^L (\omega *^k \omega - \omega \circledast^k \omega)(s) \, \dd s
	&	\leq \sum\limits_{\ell=1}^{k-1} \ee^{ \ell \alpha^+(\lambda)} \ee^{-L \lambda}
		+ \sum\limits_{\ell=1}^{k-1} \ee^{ \ell \alpha^-(\lambda)} \ee^{-L \lambda} \\ & =
		\bigg( \frac{\ee^{k \alpha^+(\lambda)}-\ee^{ \alpha^+(\lambda)}}{\ee^{\alpha^+(\lambda)}-1}
		+ \frac{\ee^{k \alpha^-(\lambda)}-\ee^{ \alpha^-(\lambda)}}{\ee^{\alpha^-(\lambda)}-1} \bigg) \, \ee^{-L \lambda}.
	\end{aligned}
\end{equation*}
\end{proof}
\end{lem}

\subsection{Proof of Theorem 10 (Total Error)} 

\textbf{Proof of Theorem 10.} Let $\alpha^+(\lambda)$ and $\alpha^-(\lambda)$ be defined as in \eqref{eq:conn}.
Combining the bound \eqref{eq:chernoff} and the bounds given by Lemmas~\ref{lem:period} and~\ref{lem:trunc}, we find that
\begin{equation*}
	\begin{aligned}
		\abs{\delta(\veps) - \widetilde{\delta}(\veps)} 
\leq	& \, \ee^{k \alpha^+(\lambda)}  \ee^{- \lambda L} + 
		\big(  \ee^{k \alpha^+(\lambda)} + \ee^{k \alpha^-(\lambda)}  \big) \frac{ \ee^{- L \lambda} }{ 1 - \ee^{- L \lambda}} \\
		& \quad \quad \quad + \bigg( \frac{\ee^{k \alpha^+(\lambda)}-\ee^{ \alpha^+(\lambda)}}{\ee^{\alpha^+(\lambda)}-1}
		+ \frac{\ee^{k \alpha^-(\lambda)}-\ee^{ \alpha^-(\lambda)}}{\ee^{\alpha^-(\lambda)}-1} \bigg) \, \ee^{-L \lambda} \\
\leq & \, \ee^{k \alpha^+(\lambda)}  \frac{ \ee^{- L \lambda} }{ 1 - \ee^{- L \lambda}} + 
		\big(  \ee^{k \alpha^+(\lambda)} + \ee^{k \alpha^-(\lambda)}  \big) \frac{ \ee^{- L \lambda} }{ 1 - \ee^{- L \lambda}} \\ 
		& \quad \quad \quad + \bigg( \frac{\ee^{k \alpha^+(\lambda)}-\ee^{ \alpha^+(\lambda)}}{\ee^{\alpha^+(\lambda)}-1}
		+ \frac{\ee^{k \alpha^-(\lambda)}-\ee^{ \alpha^-(\lambda)}}{\ee^{\alpha^-(\lambda)}-1} \bigg) \, \frac{ \ee^{- L \lambda} }{ 1 - \ee^{- L \lambda}} \\
= & \, \bigg(  \frac{2 \ee^{(k+1) \alpha^+(\lambda)} -\ee^{k \alpha^+(\lambda)}  -   \ee^{ \alpha^+(\lambda)}}{\ee^{\alpha^+(\lambda)} - 1}
		+ \frac{ \ee^{(k+1) \alpha^-(\lambda)} - \ee^{ \alpha^-(\lambda)}}{\ee^{\alpha^-(\lambda)}-1} \bigg) \, \frac{\ee^{-L \lambda}}{1- \ee^{-L \lambda}}. 
	\end{aligned}
\end{equation*}
	\qed


\section{Theorem 11: Tight Bound for Multidimensional Mechanisms via One Dimensional Distributions}

The following results shows that the tight $(\veps,\delta)$-bound for a multidimensional mechanism $\mathcal{M}$
can be obtained by analysis of one dimensional distributions, in case the neighbouring datasets $X$ and $Y$ leading to the maximal $\delta(\veps)$ are known.

\begin{thm} \label{thm:equivalence}
Consider a function $f \, : \, \mathcal{X}^N \rightarrow \mathbb{R}^d$ and a randomised mechanism
$\mathcal{M}$ of the form
$
\mathcal{M}(X) = f(X) + Z,
$
where $Z_i$'s are independent random variables. Suppose the data sets $X$ and $Y$ lead to the $\delta(\veps)$-upper bound, and denote $\Delta = f(X) - f(Y)$.
Then, the tight $(\veps,\delta)$-bound for $\mathcal{M}$ 
is given by the tight $(\veps,\delta)$-bound for the non-adaptive compositions of one-dimensional random variables
$$
\Delta_i + Z_i \quad \textrm{and} \quad Z_i, \quad 1 \leq i \leq d.
$$
\begin{proof}
The claim can be shown simply by observing that the privacy loss distribution generated by $\mathcal{M}(X)$ and $\mathcal{M}(Y)$
and the privacy loss distribution generated by compositions 
$(\Delta_1 + Z_1, \ldots, \Delta_d + Z_d)$ and $(Z_1, \ldots, Z_d)$	are the same.
\end{proof}
\end{thm}	

\section{Experiments of Section 6.2} 

We next show how to use the Fourier accountant for obtaining the $(\veps,\delta)$-bound of Figure 4. 
Essentially, we show how to obtain the PLD for a subsampled multivariate mechanism, where the neighbouring distributions are known and fixed (i.e., $\Delta = f(X) - f(Y)$ is fixed and $f(X)$ is sampled with probability $q$ and $f(Y)$ with probability $1-q$).

Now denote the density functions for one-dimensional mechanisms $\mathcal{M}(X)$ and $\mathcal{M}(Y)$ by
$$
f_X(t) := \sum_i a_{X,i} \cdot \deltat{t_{X,i}} \quad \textrm{and} \quad
f_Y(t) := \sum_i a_{Y,i} \cdot \deltat{t_{Y,i}},
$$
respectively.

Then, for the $d$-fold compositions
$$
\big(\mathcal{M}(X), \ldots, \mathcal{M}(X)\big) \quad \textrm{and} \quad
\big(\mathcal{M}(Y), \ldots, \mathcal{M}(Y)\big),
$$
the density functions are given by the convolutions
$$
\widetilde{f}_X(t) = \sum_{(i_1, \ldots, i_d)} a_{X,i_1} \cdots a_{X,i_d} \cdot \deltat{t_{X,i_1} + \ldots + t_{X,i_d} } \quad \textrm{and} \quad
\widetilde{f}_Y(t) = \sum_{(i_1, \ldots, i_d)} a_{Y,i_1} \cdots a_{Y,i_d} \cdot \deltat{t_{Y,i_1} + \ldots + t_{Y,i_d} },
$$
respectively.
%

By definition, the PLD generated by the distributions 
$$
q \cdot \widetilde{f}_X + (1-q) \cdot \widetilde{f}_Y \quad \textrm{and} \quad \widetilde{f}_Y, 
$$
is of the form
\begin{equation} \label{eq:omega_mixture}
	\widetilde{\omega}(s) = \sum_{(i_1, \ldots, i_d)}  \bigg( q \cdot a_{X,i_1} \cdots a_{X,i_d} + (1-q) \cdot a_{Y,i_1} \cdots a_{Y,i_d} \bigg) \cdot \deltas{\widetilde{s}_i},
\end{equation}
where
\begin{equation*}
	\begin{aligned}
		\widetilde{s}_i &= \log \bigg(  \frac{ q \cdot a_{X,i_1} \cdots a_{X,i_d} + (1-q) \cdot a_{Y,i_1} \cdots a_{Y,i_d} }{  a_{Y,i_1} \cdots a_{Y,i_d}  }     \bigg) \\
		&= \log \bigg(  q \cdot \frac{ a_{X,i_1}\cdots a_{X,i_d} }{  a_{Y,i_1} \cdots a_{Y,i_d}  }  + (1-q)       \bigg) \\
		&= \log \bigg(  q \cdot \exp\big( s_{i_1} + \ldots s_{i_d}   \big)  + (1-q)       \bigg),  \\
	\end{aligned}
\end{equation*}
where 
$$
s_i = \log\bigg( \frac{a_{X,i}}{a_{Y,i}}\bigg)
$$ 
for all $i$.
Thus, if we have the distributions 
\begin{equation} \label{eq:omega_term1x}
	\omega_1(s) = \sum_{(i_1, \ldots, i_d)} a_{X,i_1} \cdots a_{X,i_d} \cdot \deltas{s_{i_1} + \ldots s_{i_d}}
\end{equation}
and
\begin{equation}  \label{eq:omega_term2y}
	\omega_2(s) = \sum_{(i_1, \ldots, i_d)} a_{Y,i_1} \cdots a_{Y,i_d} \cdot \deltas{s_{i_1} + \ldots s_{i_d}},
\end{equation}
we can form the PLD $\widetilde{\omega}$ by the change of variable
$$
s \rightarrow \log \big( q \cdot s   + (1-q)       \big) 
$$
and summing the coefficients as in \eqref{eq:omega_mixture}. On the other hand, we can obtain $\omega_1$ and $\omega_2$ by using the Fourier accountant
to the $d$-fold convolutions of the distributions
$$
 \sum_i a_{X,i} \cdot \deltas{s_i} \quad \textrm{and} \quad
 \sum_i a_{Y,i} \cdot \deltas{s_i}.
$$
Also, the $\delta(\infty)$-probabilities can be evaluated straightforwardly for $q \cdot \widetilde{f}_X + (1-q) \cdot \widetilde{f}_Y$ and $\widetilde{f}_Y$.

\section{Section 6.3: The Subsampled Gaussian Mechanism} \label{subsec:subsampled_gauss}

In this Section we give an error analysis for the approximations given in Section 6.3.
Recall first the form of the PLD for the subsampled Gaussian mechanism. For a subsampling ratio $0<q<1$ and noise level $\sigma>0$, 
the continuous PLD distribution is given by
\begin{equation} \label{eq:subsampled_PLD_def} 
\omega(s) = \begin{cases}
f(g(s))g'(s), &\text{ if }  s > \log(1-q), \\
0, &\text{ otherwise},
\end{cases}
\end{equation}
where
\begin{equation} \label{eq:fg} 
f(t) = \frac{1}{\sqrt{2 \pi \sigma^2}} \, [ q \ee^{ \frac{-(t-1)^2}{2 \sigma^2}} + (1-q) \ee^{-\frac{t^2}{2 \sigma^2}} ],
\quad \quad
g(s) = \sigma^2 \log \left( \frac{\ee^s - (1-q)}{q} \right) + \frac{1}{2}.
\end{equation}

In order to carry out an error analysis for the approximations given in Section 6.3, 
we define the infinite extending grid approximations of $\omega_{\mathrm{min}}$ and $\omega_{\mathrm{max}}$.
Let $L>0$, $n \in \mathbb{Z}^+$, $\Delta x = 2L/n$ and let the grid $X_n$ be defined as in \eqref{eq:grid}.
Define
\begin{equation*} 
	\begin{aligned}
		\omega_{\mathrm{min}}(s) = \sum\limits_{i=0}^{n-1} c^-_i \cdot \deltas{s_i}, \quad \quad
		\omega_{\mathrm{max}}(s) = \sum\limits_{i=0}^{n-1} c^+_i \cdot \deltas{s_i},
	\end{aligned}
\end{equation*}
where $s_i = i \Delta x$ and 
\begin{equation} \label{eq:c_plusminus}
	\begin{aligned}
		c^-_i = \Delta x \cdot \min\limits_{s \in [s_i, s_{i+1}]} \omega(s), \quad \quad 
		c^+_i = \Delta x \cdot \max\limits_{s \in [s_{i-1}, s_i]} \omega(s).
	\end{aligned}
\end{equation}
Define
\begin{equation} \label{eq:c_plusminus_inf}
	\begin{aligned}
		\omega^\infty_{\mathrm{min}}(s) = \sum\limits_{i \in \mathbb{Z}} c^-_i \cdot \deltas{s_i}, \quad \quad 
		\omega^\infty_{\mathrm{max}}(s) = \sum\limits_{i \in \mathbb{Z}} c^+_i \cdot \deltas{s_i},
	\end{aligned}
\end{equation}
where $c^-_i$ and $c^+_i$ are as defined in \eqref{eq:c_plusminus}. We find that $\omega$ as defined in \eqref{eq:subsampled_PLD_def} 
has one stationary point which we determine numerically.
Using this, the numerical values of $c^-_i$ and $c^+_i$ are obtained.

\emph{We obtain approximations for the lower and upper bounds $\delta_{\mathrm{min}}(\veps)$ and $\delta_{\mathrm{max}}(\veps)$ of Section 6.3 by running Algorithm 1 for
$\omega^\infty_{\mathrm{min}}$ and $\omega^\infty_{\mathrm{max}}$ using some prescribed parameter values $n$ and $L$.
This is equivalent to running Algorithm 1 for the truncated distributions 
$\omega_{\mathrm{min}}$ and $\omega_{\mathrm{max}}$. However, to obtain the bounds of Theorem 10 
(and subsequently strict lower and upper bounds for $\delta(\veps)$), 
the error analysis has to be carried out for the distributions
$\omega^\infty_{\mathrm{min}}$ and $\omega^\infty_{\mathrm{max}}$. To this end, we need bounds for
 the moment generating functions of $-\omega^\infty_{\mathrm{min}}$, $\omega^\infty_{\mathrm{min}}$
 $-\omega^\infty_{\mathrm{max}}$ and $\omega^\infty_{\mathrm{max}}$.  }

However, we first show that $\omega^\infty_{\mathrm{min}}$ and $\omega^\infty_{\mathrm{max}}$ indeed give lower and upper bounds for $\delta(\veps)$.

\begin{lem} \label{lem:subsampled_ineq}
Let $\delta(\veps)$ be given by the integral formula of Lemma~\ref{lem:maxrepr} 
for some privacy loss distribution $\omega$ and for some $\delta(\infty) \geq 0$.
Let $\delta_{\mathrm{min}}^\infty(\veps)$ and $\delta_{\mathrm{max}}^\infty(\veps)$ be defined analogously by 
$\omega^\infty_{\mathrm{min}}$ and $\omega^\infty_{\mathrm{max}}$.
Then for all $\veps>0$ we have
$$
\delta_{\mathrm{min}}^\infty(\veps) \leq \delta(\veps) \leq \delta_{\mathrm{max}}^\infty(\veps).
$$
\begin{proof}
From the definition \eqref{eq:c_plusminus_inf} and from
the fact that $(1-\ee^{\veps - s})$ is a monotonously increasing function of $s$ it follows that
the discrete sums $\delta_{\mathrm{min}}^\infty(\veps)$ and $\delta_{\mathrm{max}}^\infty(\veps)$
are the lower and upper Riemann sums for the continuous integral $\delta(\veps)$ on the partition $\{ i \Delta x \, : \, i \in \mathbb{Z} \}$.
This shows the claim.
\end{proof}
\end{lem}

Lemma~\ref{lem:subsampled_ineq} directly generalises to convolutions:
\begin{cor}
Consider a single composition, i.e., suppose the PLD
is given by $\omega * \omega$ for a distribution $\omega$ of the form \eqref{eq:subsampled_PLD_def}. 
Let $\omega^\infty_{\mathrm{max}}$ be defined as in \eqref{eq:c_plusminus_inf}. 
We have that
\begin{equation} \label{eq:bound_omega_max}
	\begin{aligned}
		\int\limits_\veps^\infty  (1-\ee^{\veps - s}) \, (\omega * \omega)(s)   \, \dd s &= 
		 \int\limits_\veps^\infty  (1-\ee^{\veps - s}) \,  \int\limits_{-\infty}^\infty  \omega(t) \, \omega(s-t) \, \dd t \, \dd s \\
		 &= \int\limits_{-\infty}^\infty  \omega(t)  \int\limits_\veps^\infty  (1-\ee^{\veps - s}) \, \omega(s-t) \, \dd s   \, \dd t  \\
		 &\leq \int\limits_{-\infty}^\infty  \omega(t)  \int\limits_\veps^\infty  (1-\ee^{\veps - s}) \, \omega^\infty_{\mathrm{max}}(s-t) \, \dd s   \, \dd t  \\
		 &= \int\limits_{-\infty}^\infty  \omega(t)  \int\limits_\veps^\infty  (1-\ee^{\veps - s}) \sum\limits_{i \in \mathbb{Z}} c^+_i \cdot \deltas{s_i+t} \, \dd s   \, \dd t  \\
		 &=  \int\limits_{-\infty}^\infty  \omega(t) \sum\limits_{s_i + t > \veps} (1-\ee^{\veps - (s_i+t)}) \, c^+_i  \, \dd t \\
		 &\leq   \int\limits_{-\infty}^\infty  \omega^\infty_{\mathrm{max}}(t) \sum\limits_{s_i + t > \veps} (1-\ee^{\veps - (s_i+t)}) \, c^+_i  \, \dd t \\
		 &= \sum\limits_{s_i + s_j > \veps} (1-\ee^{\veps - (s_i+s_j)}) \, c^+_i c^+_j \\
		 &=  \int\limits_\veps^\infty  (1-\ee^{\veps - s}) \, \sum\limits_{i,j}  c^+_i c^+_j  \deltas{s_i+s_j} \, \dd s \\
		 &= \int\limits_\veps^\infty  (1-\ee^{\veps - s}) \, (\omega^\infty_{\mathrm{max}} * \omega^\infty_{\mathrm{max}})(s) \, \dd s.
	\end{aligned}
\end{equation}
Showing that 
$$
\int\limits_\veps^\infty  (1-\ee^{\veps - s}) \, (\omega * \omega)(s)  \geq \int\limits_\veps^\infty  (1-\ee^{\veps - s}) \, 
(\omega^\infty_{\mathrm{min}} * \omega^\infty_{\mathrm{min}})(s) \, \dd s
$$
goes analogously.
Inductively, bounding as in \eqref{eq:bound_omega_max}, we also see that
\begin{equation*}
	\begin{aligned}
\int\limits_\veps^\infty  (1-\ee^{\veps - s}) \, (\omega *^k \omega)(s)   \, \dd s 
&\leq \sum\limits_{s_{i_1} + \ldots + s_{i_k} > \veps} (1-\ee^{\veps - (s_{i_1} + \ldots + s_{i_k})}) \, a_{i_1} \cdot \ldots \cdot a_{i_k} \\
&= \int\limits_\veps^\infty  (1-\ee^{\veps - s}) \, (\omega^\infty_{\mathrm{max}} *^k \omega^\infty_{\mathrm{max}})(s) \, \dd s
	\end{aligned}
\end{equation*}
and similarly for the lower bound determined by the convolutions of $\omega^\infty_{\mathrm{min}}$.
\end{cor}

To evaluate $\alpha^+(\lambda)$ and $\alpha^-(\lambda)$ in the upper bound of Theorem 10 of the main text, 
we need the moment generating functions of $-\omega^\infty_{\mathrm{min}}$, $\omega^\infty_{\mathrm{min}}$, 
$-\omega^\infty_{\mathrm{max}}$ and $\omega^\infty_{\mathrm{max}}$.
We first state the following auxiliary lemma needed to bound these moment generating functions.

\begin{lem} \label{lem:boundy2}

For all $s \geq 1$ and $0<q\leq \tfrac{1}{2}$:
$$
\omega(s) \leq \sigma \sqrt{ \frac{2}{\pi} } \ee^{ - \frac{ ( \sigma^2 s + C)^2 }{2 \sigma^2} }, 
$$
where $C = \sigma^2 \log( \frac{1}{2q}) - \frac{1}{2}$.

\begin{proof}
When $s\geq1$,
\begin{equation} \label{eq:in}
\ee^s - (1-q) \geq \frac{1}{2} \ee^s	
\end{equation}
and subsequently
$$
g(s) = \sigma^2 \log \left( \frac{\ee^s - (1-q)}{q} \right) + \frac{1}{2} \geq 
\sigma^2 s + \wt{C},
$$
where $\wt{C} = \sigma^2 \log( \tfrac{1}{2q}) + \tfrac{1}{2}$. We see that when $0<q \leq \tfrac{1}{2}$, we have $g(s) \geq \tfrac{1}{2}$.
From \eqref{eq:fg} we see that 
$$
f(g(s)) \leq  \frac{1}{\sqrt{2 \pi \sigma^2}} \ee^{ - \frac{ ( \sigma^2 s + \wt{C} - 1)^2 }{2 \sigma^2} },
$$
Furthermore, when $s \geq 1$, from \eqref{eq:in} it follows that
$$
g'(s) = \frac{\sigma^2 \ee^s}{\ee^s - (1-q)}  \leq 2 \sigma^2.
$$
Thus, when $s \geq 1$,
$$
\omega(s) \leq \sigma \sqrt{ \frac{2}{\pi} } \ee^{ - \frac{( \sigma^2 s + C )^2 }{2 \sigma^2} },
$$
where $C = \sigma^2 \log( \tfrac{1}{2q}) - \tfrac{1}{2}$.
\end{proof}
\end{lem}

Using Lemma~\ref{lem:boundy2}, we can bound the moment generating function of $\omega^\infty_{\mathrm{max}}$
as follows. We note that $\mathbb{E} [\ee^{\lambda \omega_{\mathrm{max}} }]$ can be evaluated numerically.

\begin{lem} \label{lem:mgfs}
Let $0 < \lambda \leq L$ and assume $\sigma \geq 1$ and $\Delta x \leq c \cdot L$, $0<c<1$.
The moment generating function of 
$\omega^\infty_{\mathrm{max}}$ can be bounded as
$$
\mathbb{E} [\ee^{\lambda \omega^\infty_{\mathrm{max}} }] \leq 
\mathbb{E} [\ee^{\lambda \omega_{\mathrm{max}} }] + \mathrm{err}(\lambda,L,\sigma),
$$
where
\begin{equation} \label{eq:err_term}
	\mathrm{err}(\lambda,L,\sigma) = \ee^{c \lambda L } \frac{2}{\sqrt{ \pi} }  \ee^{ - \frac{ \lambda(2C - \lambda) }{2 \sigma^2} } 
			  \mathrm{erfc} \left( \frac{ (1-c) \sigma^2 L + C - \lambda }{ \sqrt{2} \sigma }   \right).
\end{equation}
Here $\omega_{\mathrm{max}}$ is the restriction of $\omega^\infty_{\mathrm{max}}$ to the interval $[-L,L]$ 
(i.e., as defined in equation (14) of the main text) 
and the constant $C$ is as defined in Lemma~\ref{lem:boundy2}.
\begin{proof}
Assuming $L > \abs{\log{1-q}}$ (i.e., $\omega(s)=0$ for all $s < -L$),
the moment generating function of $\omega^\infty_{\mathrm{max}}$ is given by
\begin{equation} \label{eq:pld_lmf}
	\begin{aligned}
		\mathbb{E} [\ee^{\lambda \omega^\infty_{\mathrm{max}} }] &= \int_{-\infty}^L \ee^{\lambda s} \omega^\infty_{\mathrm{max}}(s) \, \dd s 
		+ \int_{L}^\infty \ee^{\lambda s} \omega^\infty_{\mathrm{max}}(s) \, \dd s \\
		&= \int_{-L}^L \ee^{\lambda s} \omega^\infty_{\mathrm{max}}(s) \, \dd s 
		+ \int_{L}^\infty \ee^{\lambda s} \omega^\infty_{\mathrm{max}}(s) \, \dd s \\
		&= \mathbb{E} [\ee^{\lambda \omega_{\mathrm{max}} }]
		+  \sum\limits_{i \geq n } \Delta x \cdot \ee^{\lambda  i \Delta x } \cdot c^+_i.
	\end{aligned}
\end{equation}
From Lemma~\ref{lem:boundy2} it follows that
$$
c^+_i  = \max\limits_{s \in [s_{i-1}, s_i]} \omega(s) 
\leq \sigma \sqrt{ \frac{2}{\pi} } \ee^{ - \frac{ ( \sigma^2 s_{i-1} + C)^2 }{2 \sigma^2} },
$$
where $C = \sigma^2 \log( \frac{1}{2q}) - \frac{1}{2}$, $s_i=i \Delta x$.
Thus 
\begin{equation}
	\begin{aligned}
		\sum\limits_{i \geq n }  \ee^{\lambda  i \Delta x } \cdot c^+_i &  = 
		\ee^{\lambda \Delta x } \sum\limits_{i \geq n } \ee^{\lambda  s_{i-1} } \cdot c^+_i \\		
		&\leq 
		\ee^{\lambda \Delta x } \sigma \sqrt{ \frac{2}{\pi} } 
		\sum\limits_{i \geq n } \Delta x \cdot \ee^{\lambda  s_{i-1} }
		 \ee^{ - \frac{ ( \sigma^2 s_{i-1} + C)^2 }{2 \sigma^2} } \\
		& = \ee^{\lambda \Delta x } \sigma \sqrt{ \frac{2}{\pi} }  
		\sum\limits_{i \geq n } \Delta x \cdot \ee^{  \frac{ - ( \sigma^2 s_{i-1} + C - \lambda)^2 - \lambda(2C - \lambda) }{2 \sigma^2} } \\
		& = \ee^{\lambda \Delta x } \sigma \sqrt{ \frac{2}{\pi} }  \ee^{ - \frac{ \lambda(2C - \lambda) }{2 \sigma^2} }
		\sum\limits_{i \geq n } \Delta x \cdot \ee^{ - \frac{ ( \sigma^2 s_{i-1} + C - \lambda)^2 }{2 \sigma^2} }.
	\end{aligned}
\end{equation}
Assuming $\sigma \geq 1$ and $\lambda \leq L$, $\Delta x \leq c \cdot L$, we further see that 
\begin{equation}
	\begin{aligned}
		& \ee^{\lambda \Delta x } \sigma \sqrt{ \frac{2}{\pi} }  \ee^{ - \frac{ \lambda(2C - \lambda) }{2 \sigma^2} }
		\sum\limits_{i \geq n } \Delta x \cdot \ee^{ - \frac{ ( \sigma^2 s_{i-1} + C - \lambda)^2 }{2 \sigma^2} }  \\
		& \leq \ee^{\lambda \Delta x } \sigma \sqrt{ \frac{2}{\pi} }  \ee^{ - \frac{ \lambda(2C - \lambda) }{2 \sigma^2} } 
		\int\limits_{L-\Delta x}^\infty \ee^{ - \frac{ ( \sigma^2 s + C - \lambda)^2 }{2 \sigma^2} } \, \dd s \\
		& \leq \ee^{c \lambda L } \sigma \sqrt{ \frac{2}{\pi} }  \ee^{ - \frac{ \lambda(2C - \lambda) }{2 \sigma^2} } 
		\int\limits_{(1-c) L}^\infty \ee^{ - \frac{ ( \sigma^2 s + C - \lambda)^2 }{2 \sigma^2} } \, \dd s \\
		& = \ee^{c \lambda L } \sigma \sqrt{ \frac{2}{\pi} }  \ee^{ - \frac{ \lambda(2C - \lambda) }{2 \sigma^2} } 
		\frac{\sqrt{2}}{\sigma}  \mathrm{erfc} \left( \frac{ (1-c) \sigma^2 L + C - \lambda }{ \sqrt{2} \sigma }   \right) \\
		& = \ee^{c \lambda L } \frac{2}{\sqrt{ \pi} }  \ee^{ - \frac{ \lambda(2C - \lambda) }{2 \sigma^2} } 
		  \mathrm{erfc} \left( \frac{ (1-c) \sigma^2 L + C - \lambda }{ \sqrt{2} \sigma }   \right). \\
	\end{aligned}
\end{equation}
\end{proof}
\end{lem}
Using a reasoning similar to the proof of Lemma~\ref{lem:mgfs}, we get the following. 
We note that $\mathbb{E} [\ee^{ - \lambda \omega_{\mathrm{max}} }]$,
$\mathbb{E} [\ee^{\lambda \omega_{\mathrm{min}} }]$ and
$\mathbb{E} [\ee^{ - \lambda \omega_{\mathrm{min}} }]$ can be evaluated numerically.
\begin{cor}
The moment generating functions of $-\omega^\infty_{\mathrm{max}}$, $\omega^\infty_{\mathrm{min}}$ and $-\omega^\infty_{\mathrm{min}}$ can be bounded as
$$
\mathbb{E} [\ee^{ - \lambda \omega^\infty_{\mathrm{max}} }] \leq 
\mathbb{E} [\ee^{ - \lambda \omega_{\mathrm{max}} }] + \mathrm{err}(\lambda,L,\sigma),
$$
$$
\mathbb{E} [\ee^{\lambda \omega^\infty_{\mathrm{min}} }] \leq 
\mathbb{E} [\ee^{ \lambda \omega_{\mathrm{min}} }] + \mathrm{err}(\lambda,L,\sigma),
$$
$$
\mathbb{E} [\ee^{ - \lambda \omega^\infty_{\mathrm{min}} }] \leq 
\mathbb{E} [\ee^{ - \lambda \omega_{\mathrm{min}} }] + \mathrm{err}(\lambda,L,\sigma),
$$
where
$\mathrm{err}(\lambda,L,\sigma)$ is defined as in \eqref{eq:err_term}.
\begin{proof}
%
Assuming $L > \abs{\log{1-q}}$ (i.e., $\omega(s)=0$ for all $s < -L$),
the moment generating function of $-\omega^\infty_{\mathrm{max}}$ is given by
\begin{equation} \label{eq:pld_lmf2}
	\begin{aligned}
		\mathbb{E} [\ee^{ - \lambda \omega^\infty_{\mathrm{max}} }] &= \int\limits_{-\infty}^L \ee^{- \lambda s} \omega^\infty_{\mathrm{max}}(s) \, \dd s 
		+ \int\limits_{L}^\infty \ee^{-\lambda s} \omega^\infty_{\mathrm{max}}(s) \, \dd s \\
		&= \int\limits_{-L}^L \ee^{-\lambda s} \omega^\infty_{\mathrm{max}}(s) \, \dd s 
		+ \int\limits_{L}^\infty \ee^{- \lambda s} \omega^\infty_{\mathrm{max}}(s) \, \dd s \\
		& \leq \int\limits_{-L}^L \ee^{-\lambda s} \omega^\infty_{\mathrm{max}}(s) \, \dd s 
		+ \int\limits_{L}^\infty \ee^{ \lambda s} \omega^\infty_{\mathrm{max}}(s) \, \dd s.
	\end{aligned}
\end{equation}
After bounding the term $\int_{L}^\infty \ee^{ \lambda s} \omega^\infty_{\mathrm{max}}(s) \, \dd s$ as in the proof of Lemma~\ref{lem:mgfs}, the
first claim follows. Bounding $\mathbb{E} [\ee^{\lambda \omega^\infty_{\mathrm{min}} }]$ and
$\mathbb{E} [\ee^{ - \lambda \omega^\infty_{\mathrm{min}} }]$ can be carried out analogously to \eqref{eq:pld_lmf2}.
\end{proof}
\end{cor}

\begin{remark}
In the experiments, the effect of the error term $\mathrm{err}(\lambda,L,\sigma)$ was found to be negligible
(less than $10^{-90}$ in the experiments of Figure 3).
\end{remark}


\section{Description of Learning Rate Cooling Used for Experiments of Figure 2b.}

When running the feedforward network experiment of Figure 2b, we set the initial learning rate $\eta=0.02$.
When $n=2400$ and $\abs{B}=500$, starting from epoch 13, and when $n=3000$ and $\abs{B}=300$, starting from epoch 5,
the learning rate $\eta$ is linearly decreased after each epoch such that it is zero at the end of the training.

\end{document}